\documentclass[10pt,twocolumn,letterpaper]{article}

\usepackage{iccv}
\usepackage{times}
\usepackage{epsfig}
\usepackage{graphicx}
\usepackage{amsmath}
\usepackage{amssymb}

\usepackage{booktabs}
\usepackage[skip=4pt]{caption}
\usepackage{adjustbox}
\usepackage{multirow}
\usepackage{balance}
\usepackage{microtype}
\usepackage{xcolor}
\usepackage{enumitem}
\usepackage{cuted}
\usepackage{capt-of}
\usepackage[accsupp]{axessibility}  %

\usepackage[linesnumbered,ruled,vlined]{algorithm2e}
\SetKwInput{KwInput}{Input}                %
\SetKwInput{KwOutput}{Output}              %
\usepackage{mathtools}

\usepackage[numbers,sort,compress]{natbib}
\usepackage{lipsum}

\usepackage[pagebackref=true,breaklinks=true,letterpaper=true,colorlinks,bookmarks=false]{hyperref}

\usepackage{lipsum}

\newcommand\blfootnote[1]{%
  \begingroup
  \renewcommand\thefootnote{}\footnote{#1}%
  \addtocounter{footnote}{-1}%
  \endgroup
}

\iccvfinalcopy %

\def\iccvPaperID{1494} %

\ificcvfinal\fi

\begin{document}

\title{Discover the Unknown Biased Attribute of an Image Classifier}

\author{Zhiheng Li \hspace{10mm} Chenliang Xu\\
University of Rochester\\
{\tt\small \{zhiheng.li,chenliang.xu\}@rochester.edu}
}

\maketitle
\ificcvfinal\thispagestyle{empty}\fi

\begin{abstract}
Recent works find that AI algorithms learn biases from data. Therefore, it is urgent and vital to identify biases in AI algorithms. However, the previous bias identification pipeline overly relies on human experts to conjecture potential biases (e.g., gender), which may neglect other underlying biases not realized by humans. To help human experts better find the AI algorithms' biases, we study a new problem in this work -- for a classifier that predicts a target attribute of the input image, discover its unknown biased attribute.

To solve this challenging problem, we use a hyperplane in the generative model's latent space to represent an image attribute; thus, the original problem is transformed to optimizing the hyperplane's normal vector and offset. We propose a novel total-variation loss within this framework as the objective function and a new orthogonalization penalty as a constraint. The latter prevents trivial solutions in which the discovered biased attribute is identical with the target or one of the known-biased attributes. Extensive experiments on both disentanglement datasets and real-world datasets show that our method can discover biased attributes and achieve better disentanglement w.r.t. target attributes. Furthermore, the qualitative results show that our method can discover unnoticeable biased attributes for various object and scene classifiers, proving our method's generalizability for detecting biased attributes in diverse domains of images.
\end{abstract}

\begin{figure}
   \centering
   \includegraphics[width=0.95\linewidth]{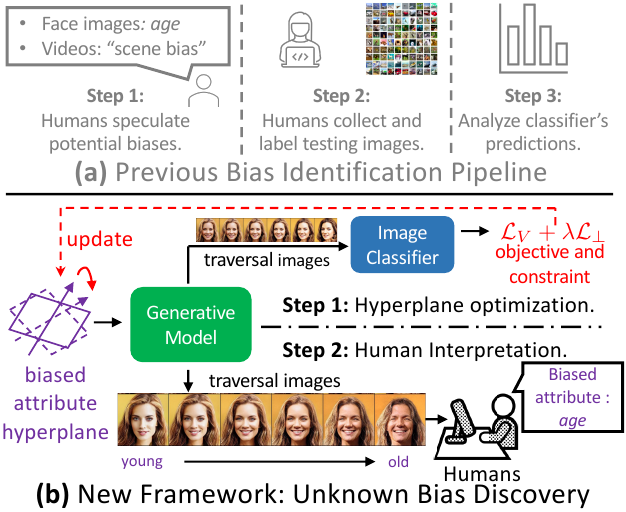}
   \caption{
      While the previous pipeline \textcolor{gray}{\textbf{(a)}} overly relies on human efforts to identify biases, we devise a new automated framework \textbf{(b)} that helps humans discover the unknown biased attribute of an image classifier. In step 1, a generative models' biased attribute hyperplane in the latent space is optimized by our proposed optimization objective and constraints. In step 2, humans can interpret the semantic meaning of the biased attribute hyperplane from the transformation in the synthesized traversal images. For example, images changing from ``young'' to ``old'' imply the biased attribute \textit{age}.
   }
   \label{fig:teaser}
   \vspace{-4mm}
\end{figure}

\section{Introduction}
\blfootnote{The code is available at \url{https://git.io/J3kMh}.} Although the performance of deep neural networks is greatly improved by training on large-scale datasets, worrisome biases are also learned by AI algorithms. Thus it is imperative to identify AI algorithms’ biases, whereas the previous bias identification pipeline~\cite{buolamwini2018ACMConf.FairnessAccount.Transpar.,mitchell2019ACMConf.FairnessAccount.Transpar.} has some shortcomings. First, it overly relies on human experts to speculate potential biases (Step 1 in Fig.~\ref{fig:teaser} (a)), which may leave other unconsidered biases unexposed. For example, people may conjecture legally protected attributes (\eg, \textit{age}, \textit{gender}\footnote{In this paper, we use \textit{gender} to denote visually perceived gender, which does not indicate the person's true gender identity.}) for face image classifiers and consider ``scene bias'' for the action recognition task~\cite{choi2019Adv.NeuralInf.Process.Syst.}. However, people may neglect other unnoticeable biased attributes such as \textit{hair length}~\cite{balakrishnan2020Eur.Conf.Comput.Vis.ECCV} and the \textit{presence of children}~\cite{wang2019IEEEInt.Conf.Comput.Vis.ICCVa} spuriously correlated with \textit{gender}. Although they are not legally protected attributes, not considering these biased attributes may still lead to unfairness against different genders~\cite{wang2019IEEEInt.Conf.Comput.Vis.ICCVa}. Second, the previous pipeline also needs expensive human efforts to collect testing images and annotate biased attributes (Step 2 in Fig.~\ref{fig:teaser} (a)) for analyzing classifier's predictions. When one wants to analyze biases in a new domain of images (\eg, object and scene categories in ImageNet~\cite{deng2009IEEEConf.Comput.Vis.PatternRecognit.CVPR} and Place365~\cite{zhou2018IEEETrans.PatternAnal.Mach.Intell.a}), massive human efforts of image collection and labeling are needed for each new image domain, which is not scalable. In addition, as a down-stream task of bias identification, many de-biasing methods~\cite{wang2020IEEEConf.Comput.Vis.PatternRecognit.CVPRe,creager2019Int.Conf.Mach.Learn.,sarhan2020Eur.Conf.Comput.Vis.ECCV,choi2019Adv.NeuralInf.Process.Syst.} also require well-defined biased attributes and annotations as inputs and supervisions to mitigate corresponding biases. As a result, if the previous pipeline does not identify the biases due to either negligence of biases or limited annotation budget, the biases will not be mitigated by those de-biasing methods. Therefore, it is urgent to discover unknown AI biases with less human effort.

To this end, we study a novel problem by defining the \textit{unknown biased attribute discovery task}: for a classifier that predicts a target attribute of the input image, discover its \textit{unknown} biased attribute. The ``target attribute'' stands for the attribute for prediction.
The ``biased attribute'' means an attribute that violates the fairness criteria~\cite{hardt2016Adv.NeuralInf.Process.Syst.,kusner2017Adv.NeuralInf.Process.Syst.} and differs from the target attribute. For example, if a gender classifier has different predictions over female images of different skin colors, then the \textit{skin tone} attribute is the biased attribute. The ``\textit{unknown}'' has two levels of meanings. First, it indicates that the biased attribute that is expected to be discovered is \textit{not} presumed by humans, not to mention annotating images with the biased attribute labels.
That is, the traditional pipeline in Fig.~\ref{fig:teaser} (a) does \textit{not} meet the requirement of ``\textit{unknown}.'' Second, ``\textit{unknown}'' also implies that human experts may have already known some biased attributes and expect a different one. After completing this task, the discovered biased attributes can be used as inputs for other down-stream tasks, such as algorithmic de-biasing.

We propose a novel framework (Fig.~\ref{fig:teaser} (b)) for this new task by solving two challenges. The first difficulty is how to represent and learn the ``attribute'' without any presumptions or labels. To tackle this problem, we base our method on some findings in \cite{karras2019IEEEConf.Comput.Vis.PatternRecognit.CVPR,shen2020IEEEConf.Comput.Vis.PatternRecognit.CVPR,shen2020IEEETrans.PatternAnal.Mach.Intell.} that the hyperplane in the latent space of generative models can linearly separate an attribute's values. Hence, we represent the unknown biased attribute as an optimizable hyperplane in a generative model's latent space (see ``biased attribute hyperplane'' in Fig.~\ref{fig:teaser} (b)). Different from previous methods~\cite{karras2019IEEEConf.Comput.Vis.PatternRecognit.CVPR,shen2020IEEEConf.Comput.Vis.PatternRecognit.CVPR,shen2020IEEETrans.PatternAnal.Mach.Intell.} using labels of attribute as the supervision, we propose \textit{total variation loss} ($\mathcal{L}_V$ in Fig.~\ref{fig:teaser} (b)) to optimize the hyperplane that induces the violation of the fairness criterion, without requiring any attribute labels. The second challenge is how to ensure the discovered biased attribute is different from the known ones. We propose \textit{orthogonalization penalty} ($\mathcal{L}_\perp$ in Fig.~\ref{fig:teaser} (b)) to encourage disentanglement between the biased attribute and known attributes. We also use $\mathcal{L}_\perp$ to prevent the biased attribute from being identical with the classifier's target attribute. Finally, to enable humans to interpret the semantic meaning of the optimized hyperplane, a sequence of images, dubbed as traversal images, are generated based on the optimized hyperplane. The variation along the traversal images is the semantic meaning of the optimization result. As shown in Fig.~\ref{fig:teaser}, the synthesized traversal face images gradually transform from ``young'' to ``old,'' indicating that the biased attribute found by our method is \textit{age}. In summary, compared with the previous pipeline (Fig.~\ref{fig:teaser} (a)), our framework first lets the optimization actively find the biased attribute (Step 1 in Fig.~\ref{fig:teaser} (b)) and postpones human involvement to the final step (Step 2 in Fig.~\ref{fig:teaser} (b)), which not only automatically discovers the unknown biases that human may not realize, but also exempts human efforts from annotating biased attributes on testing images.

We conduct three experiments to verify the effectiveness of our method. In the first experiment, two disentanglement datasets~\cite{lecun2004IEEEConf.Comput.Vis.PatternRecognit.CVPRa,higgins2017Int.Conf.Learn.Represent.} are used for creating large-scale experimental settings for evaluation. In the second experiment, we conduct experiments on two face datasets~\cite{liu2015IEEEInt.Conf.Comput.Vis.ICCV,karras2019IEEEConf.Comput.Vis.PatternRecognit.CVPR} for discovering biased attributes in face attribute classifiers. The first two experiments show that our method can correctly discover the biased attribute. In the third experiment, we apply our method for discovering the biased attribute in other domains of images, such as objects and scenes. The qualitative results and the user study show that our method can discover unnoticeable biases from classifiers pretrained on ImageNet~\cite{deng2009IEEEConf.Comput.Vis.PatternRecognit.CVPR} and Place365~\cite{zhou2018IEEETrans.PatternAnal.Mach.Intell.a}, proving our method's generalizability for finding biases in various image domains.

The contributions of this work are as follows. First, we propose a novel \textit{unknown biased attribute discovery task} for discovering unknown biases from classifiers. Solving the problems in this task can help humans better identify classifiers' biases. Second, we propose a novel method for this new task by optimizing the \textit{total variation loss} and the \textit{orthogonalization penalty} without any presumptions or labels of biased attributes. Lastly, we design comprehensive experiment settings and evaluation metrics to verify the effectiveness of our method, which can also be used as benchmarks for future works. Furthermore, many related fields can be benefited from our new framework for discovering unknown biases, such as algorithmic de-biasing, dataset audition, \etc (more discussions in Appendix~\ref{subsec.related_methods_areas}).

\section{Related Work}
\noindent \textbf{Bias Identification}
The previous bias identification pipeline mainly focuses on collecting testing images and analyzing the performances in different subgroups based on biased attribute value. \citet{buolamwini2018ACMConf.FairnessAccount.Transpar.} collect in-the-wild face images to analyze the error rates discrepancies in intersectional subgroups. \citet{kortylewski2018Proc.IEEEConf.Comput.Vis.PatternRecognit.Workshop,kortylewski2019Proc.IEEECVFConf.Comput.Vis.PatternRecognit.Workshop} use 3DMM~\cite{blanz1999Proc.26thAnnu.Conf.Comput.Graph.Interact.Tech.} to synthesize 3D face images in different poses and lighting conditions. \citet{muthukumar2018ArXiv181200099CsStat} alter facial attributes of face images via image processing techniques such as color theoretic methods and image cropping. \citet{denton2019IEEEConf.Comput.Vis.PatternRecognit.CVPRWorkshop,denton2019IEEEConf.Comput.Vis.PatternRecognit.CVPRWorkshopa} use PGGAN~\cite{karras2018Int.Conf.Learn.Represent.} to synthesize images with different values of attributes in CelebA dataset~\cite{liu2015IEEEInt.Conf.Comput.Vis.ICCV}. To further reduce the correlation between attributes, \citet{balakrishnan2020Eur.Conf.Comput.Vis.ECCV} additionally annotate attributes of synthesized image to generate multi-dimensional ``transects'' based on StyleGAN2~\cite{karras2020IEEEConf.Comput.Vis.PatternRecognit.CVPR}, where each dimension only changes the value of one attribute and remains other attributes unchanged. To reduce the dependency on the human labels, \citet{donderici2020ArXiv200209818CsEess} use a physics engine to synthesize face images by controlling different facial attributes values then augment the images to the real-world domain. Without relying on a specific algorithm for bias analysis, \citet{wang2020Eur.Conf.Comput.Vis.ECCVb} propose ``REVISE,'' a tool for computing datasets' statistics in terms of object, gender, and geography. \citet{manjunatha2019IEEEConf.Comput.Vis.PatternRecognit.CVPR} analyze biases of VQA~\cite{antol2015IEEEInt.Conf.Comput.Vis.ICCV} task by running rule mining algorithms. All of the methods above can only analyze the algorithm's biases from a presumed set of attributes or annotations. In contrast, we study a novel problem on discovering the \textit{unknown} biased attribute.

\noindent \textbf{Bias Mitigation} Many methods have been proposed to mitigate AI algorithms’ biases, and most of them require the supervision of protected attributes. \citet{wang2020IEEEConf.Comput.Vis.PatternRecognit.CVPRe} benchmark previous bias mitigation methods with full-supervision of protected attributes. \citet{creager2019Int.Conf.Mach.Learn.} train a VAE-based disentanglement method from protected attributes' labels so that the learned representation can be flexibly fair to multiple protected attributes during the testing time. \citet{sarhan2020Eur.Conf.Comput.Vis.ECCV} propose a method to maximize the entropy of the protected attribute's prediction and orthogonalize the mean vectors of normal distributions of target attribute and protected attributes. \citet{vowels2020IEEEConf.Comput.Vis.PatternRecognit.CVPR} propose a weakly-supervised bias mitigation method, NestedVAE, which is trained with paired images from different protected attribute values. \citet{choi2020Int.Conf.Mach.Learn.} use the weak supervision from a small reference dataset with balanced distribution to mitigate the biases. Though different levels of supervision are used, all of the works mentioned above require the selected protected attributes as inputs. The only exception is \cite{locatello2019Adv.NeuralInf.Process.Syst.}, where neither definition of the protected attribute nor the labels are required, and they prove that better disentanglement can decrease the unfairness score. However, the experiments in \cite{locatello2019Adv.NeuralInf.Process.Syst.} are only based on synthetic datasets with balanced distribution. In comparison, our method is tested to be effective on real-world datasets.

\noindent \textbf{Unsupervised Disentanglement}
The disentanglement methods aim to recover different independent attributes (\ie, factors of variations) from data by learning a generative model. We relate this field to our work because unsupervised disentanglement methods, which factorize attributes of data without any definitions or labels, can be used as baseline methods for the \textit{unknown biased attribute discovery task}. For VAE~\cite{kingma2014Int.Conf.Learn.Represent.}-based generated models, many methods, including \(\beta\)-VAE~\cite{higgins2017Int.Conf.Learn.Represent.}, FactorVAE~\cite{kim2018Int.Conf.Mach.Learn.}, \(\beta\)-TCVAE~\cite{chen2018Adv.NeuralInf.Process.Syst.a}, DIP-VAE-I and DIP-VAE-II~\cite{kumar2018Int.Conf.Learn.Represent.} JointVAE~\cite{dupont2018Adv.NeuralInf.Process.Syst.}, are proposed for unsupervised disentanglement, where the goal is to represent each attribute as one dimension of the hidden space of VAE. For GAN-based generative models, \citet{voynov2020Int.Conf.Mach.Learn.} train an additional reconstructor to predict the direction index and shift magnitude. More recently, Hessian Penalty~\cite{peebles2020Eur.Conf.Comput.Vis.ECCV} disentangles factors of variations by penalizing the off-diagonal items in the hessian matrix \wrt latent code. We use these unsupervised disentanglement methods as baselines to investigate their performances on the \textit{unknown biased attribute discovery task}.

\begin{figure*}[t]
   \centering
   \includegraphics[width=0.9\linewidth]{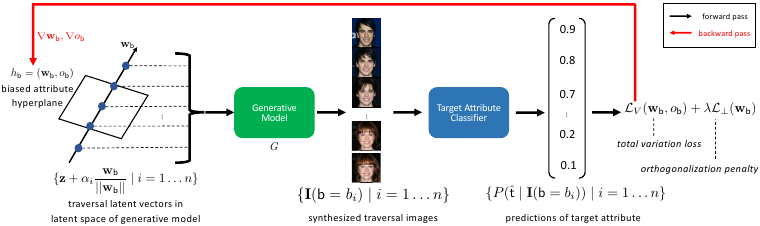}
   \caption{Method overview. We use a latent space hyperplane to represent the biased attribute. During training, we first sample the traversal latent vectors along the normal vector $\mathbf{w}_\mathsf{b}$ of the biased attribute hyperplane $h_\mathsf{b}$ in the latent space of the generative model $G$ to synthesize the traversal images, which are then are fed into the target attribute classifier. Finally, we use classifier's predictions for all traversal images to compute \textit{total variation loss} ($\mathcal{L}_V$), which is jointly minimized with \textit{orthogonalization penalty} ($\mathcal{L}_\perp$) for optimizing the hyperplane $h_\mathsf{b}$. The weights of the generative model and the classifier are fixed.}
   \label{fig.method_overview}
   \vspace{-4mm}
\end{figure*}

\section{Unknown Biased Attribute Discovery Task}
In this section, we initially introduce the definition of fairness in Sec.~\ref{subsec.fairness_def}. Then we formally define the \textit{unknown biased attribute discovery task} in Sec.~\ref{subsec.formulation_bias_attr}.

\subsection{Fairness Definition}
\label{subsec.fairness_def}

In this work, we focus on the counterfactual fairness criterion~\cite{kusner2017Adv.NeuralInf.Process.Syst.,joo2020Proc.2ndInt.WorkshopFairnessAccount.Transpar.EthicsMultimed.,denton2019IEEEConf.Comput.Vis.PatternRecognit.CVPRWorkshop,denton2019IEEEConf.Comput.Vis.PatternRecognit.CVPRWorkshopa} in the image domain and we leave studies on other fairness criteria in future works. The counterfactual fairness is formulated by:
\begin{equation}
   \mathit{P}(\hat{\mathsf{t}} \mid \mathbf{I}(\mathsf{s} = s_1) ) = \mathit{P}(\hat{\mathsf{t}} \mid \mathbf{I}(\mathsf{s} = s_2)  ), s_1 \neq s_2,
\end{equation}
where $\mathsf{t}$ is the target attribute. $s_1$ and $s_2$ are different values of the protected attribute (also called sensitive attribute) $\mathsf{s}$. $\mathbf{I}(\mathsf{s} = s_1)$ and $\mathbf{I}(\mathsf{s} = s_2)$ are pair of counterfactual images intervened in terms of the protected attribute $\mathsf{s}$ by assigning different protected attribute values: $s_1$ and $s_2$. The values of all other attributes, including the target attribute, are the same between two images. $\mathit{P}(\hat{\mathsf{t}} \mid \mathbf{I})$ is a classifier's prediction of the target attribute $\mathsf{t}$ of the image $\mathbf{I}$. For example, if $s_1$ and $s_2$ are ``young'' and ``old'' when the protected attribute is \textit{age} and the target attribute is \textit{gender}, then $\mathbf{I}(\mathsf{s} = s_1)$ and $\mathbf{I}(\mathsf{s} = s_2)$ are two images of the same person in different ages. The counterfactual fairness criterion requires the classifier's \textit{gender} prediction to be identical between two images.

\subsection{Formulation of Bias Attribute Discovery Task}
\label{subsec.formulation_bias_attr}

Here, we formally define the \textit{unknown biased attribute discovery task}. The input of this task is a classifier for predicting a target attribute $\mathsf{t}$ of the input image $\mathbf{I}$ (\ie, $\mathit{P}(\hat{\mathsf{t}} \mid \mathbf{I})$). At the same time, the classifier also learns unknown biases from its training data. We formulate such bias in the classifier as the unknown biased attribute $\mathsf{b}$ that \textbf{violates} the fairness criterion:
\begin{equation}
   \mathit{P}(\hat{\mathsf{t}} \mid \mathbf{I}(\mathsf{b} = b_1)) \neq \mathit{P}(\hat{\mathsf{t}} \mid \mathbf{I}(\mathsf{b} = b_2) ), b_1 \neq b_2,
   \label{eq.violation_parity}
\end{equation}
where $b_1$ and $b_2$ are different values of the biased attribute. Eq.~\ref{eq.violation_parity} means that the predictions of the target attribute are correlated the biased attribute. The expected output of this task is the biased attribute $\mathsf{b}$. In other words, the unknown biased attribute $\mathsf{b}$ should be discovered. Additionally, a set of known attributes $K = \{\mathsf{k}\}$ can be provided for requiring that the discovered biased attribute should not be one of known attributes. The set of known attributes is useful when a user have already known some biases that the classifier has and expect to know other unknown biased attributes of the classifier. Providing the known attribute is optional (\ie, $K = \varnothing$) when the user has not known any biased attributes.

\section{Method}
\label{sec.method}
In this section, we present our method on \textit{unknown biased attribute discovery task}. The overview of our method is shown in Fig.~\ref{fig.method_overview}. First, we represent attributes of images as the hyperplanes in the latent space of the generative model (see the left side of Fig.~\ref{fig.method_overview}). Then, we formalize our method as an optimization problem and propose \textit{total variation loss} to optimize the hyperplane of the biased attribute (see Sec.~\ref{subsec.tv_loss}). To avoid some unwanted results, we propose \textit{orthogonalization penalty} in Sec.~\ref{subsec.orth_penalty} as the constraints for the optimization problem. Finally, we summarize the full model in Sec.~\ref{subsec.full_model}.

\subsection{Representation of the Biased Attribute}
\label{subsec.represent_attr}

As defined in Sec.~\ref{subsec.formulation_bias_attr}, an unknown biased attribute $\mathsf{b}$ needs to be discovered in \textit{unknown biased attribute discovery task}. We approach this task by solving an optimization problem.
Therefore, we need to formulate the biased attribute as an optimizable representation. To this end, we leverage a generative model $\mathit{G}$ (Generative Model in Fig.~\ref{fig.method_overview}) that synthesizes the image $\mathbf{I}$ from a latent vector $\mathbf{z} \in \mathbb{R}^d$, where $d$ denotes the dimensions of $\mathit{G}$'s latent space. Here $\mathit{G}$ can be implemented by the generator of GAN~\cite{goodfellow2014Adv.NeuralInf.Process.Syst.} or the decoder of VAE~\cite{kingma2014Int.Conf.Learn.Represent.}. Some recent works in the field of image editing~\cite{shen2020IEEEConf.Comput.Vis.PatternRecognit.CVPR,shen2020IEEETrans.PatternAnal.Mach.Intell.} find that the hyperplane in generative model's latent space can be learned with full-supervision of attribute labels to linearly separate attribute values. Based on this finding, we represent the biased attribute as the hyperplane $h_\mathsf{b} = (\mathbf{w}_\mathsf{b}, o_\mathsf{b})$ in $\mathit{G}$'s latent space (left side of Fig.~\ref{fig.method_overview}), where $\mathbf{w}_\mathsf{b} \in \mathbb{R}^d$ and $o_\mathsf{b} \in \mathbb{R}$ are normal vector and offset of $\mathsf{a}$'s hyperplane. In this way, the problem of discovering the biased attribute can be transformed into an optimization problem by learning the hyperplane $h_\mathsf{b}$.

\subsection{Total Variation Loss}
\label{subsec.tv_loss}
After formalizing the biased attribute as an optimizable representation, the next question is how to design an optimization objective. Note that, different from image editing task~\cite{shen2020IEEEConf.Comput.Vis.PatternRecognit.CVPR,shen2020IEEETrans.PatternAnal.Mach.Intell.} where attribute labels are available as full supervision, we do not have labels for learning the hyperplane because the biased attribute is even unknown, not to mention collecting labels for supervised training.

To solve this challenging problem, we utilize the definition in Sec.~\ref{subsec.formulation_bias_attr} that the unknown biased attribute violates the fairness criterion. In order to check if the hyperplane $h_\mathsf{b}$ violates the fairness criterion, we generate $N$ images that have different values of the biased attribute, formulated as $\{\mathbf{I}(\mathsf{b} = b_i) \mid i = 1 \dots N\}$ (shown in the middle of Fig.~\ref{fig.method_overview}). We term these images as traversal images. We achieve this by the following steps. First, we randomly sample a latent vector $\mathbf{z}$. Then, since the normal vector $\mathbf{w}_\mathsf{b}$ of the hyperplane $h_\mathsf{b}$ is the most discriminative direction to separate different values of $\mathsf{b}$, we traverse along the normal vector starting from latent vector $\mathbf{z}$, resulting in traversal latent vectors $ \{\mathbf{z} + \alpha_i \frac{\mathbf{w}_\mathbf{b}}{||\mathbf{w}_\mathbf{b}||} \mid i = 1 \ldots N \} $ (illustrated as blue dots in Fig.~\ref{fig.method_overview}), where $\alpha_i$ is the $i$-the step size of the traversal. Finally, the traversal images can be synthesized by feeding the traversal latent vectors into the generative model $\mathit{G}$ (\ie, $\mathbf{I}(\mathsf{b} = b_i) = \mathit{G} (\mathbf{z} + \alpha_i \frac{ \mathbf{w}_b }{ ||\mathbf{w}_b||}) $).

After synthesizing the traversal images, we feed them to the classifier (``Target Attribute Classifier'' in Fig.~\ref{fig.method_overview}) and obtain the target attribute predictions of the traversal images $\{ \mathit{P}(\hat{\mathsf{t}} \mid \mathbf{I}(\mathsf{b} = b_i) ) \mid 1 \ldots N \}$. Then we propose the \textit{total variation loss} ($\mathcal{L}_V$) as the objective function, which quantifies the degree of violation against the fairness definitions:
\begin{align}
   \begin{split}
       \mathcal{L}_V = -\log \frac{1}{N-1}  \sum_{i=1}^{N-1} | & \mathit{P}(\hat{\mathsf{t}} \mid \mathit{G}(\mathbf{z} + \alpha_{i+1} \frac{\mathbf{w}_\mathsf{b}}{||\mathbf{w}_\mathsf{b}||})) \\  - & \mathit{P}(\hat{\mathsf{t}} \mid \mathit{G}(\mathbf{z} + \alpha_i \frac{\mathbf{w}_\mathsf{b}}{||\mathbf{w}_\mathsf{b}||}))|.
   \end{split}
   \label{eq.total_variation_loss}
\end{align}
Intuitively, \textit{total variation loss} checks the fairness definition of each consecutive predictions in $\{ \mathit{P}(\hat{\mathsf{t}} \mid \mathbf{I}(\mathsf{b} = b_i) ) \mid 1 \ldots N \}$, and larger differences over the predictions lead to lower \textit{total variation loss}. Since all aforementioned operations are differentiable, we minimize the \textit{total variation loss} by updating the hyperplane via gradient decent. In practice, offset $o_\mathsf{b}$ is used for projecting the sampled latent vector $\mathbf{z}$ to the hyperplane, so we optimize both $\mathbf{w}_\mathsf{b}$ and $o_\mathsf{o}$ by $\mathcal{L}_V$. The complete algorithm for computing $\mathcal{L}_V$ is in Appendix~\ref{subsec.project_code_to_plane}.

\subsection{Orthogonalization Penalty}
\label{subsec.orth_penalty}

However, minimizing \textit{total variation loss} ($\mathcal{L}_V$) alone has two problems. First, it may lead to a trivial solution where the discovered biased attribute is just the target attribute because images with different target attribute values will definitely produce large variations in the target attribute predictions. For example, a \textit{gender} classifier will have large prediction variations when the traversal images transform from ``male'' to ``female.'' Secondly, the task allows users to provide a set of known attribute $K = \{\mathsf{k}\}$ and the discovered biased attribute should not be one of the known attribute $\mathsf{k}$ (see Sec.~\ref{subsec.formulation_bias_attr}), which cannot be achieved by minimizing $\mathcal{L}_V$. To tackle these two problems, we also represent the target attribute $\mathsf{t}$ and known attribute $\mathsf{k}$ as hyperplanes in the latent space, denoted by $h_\mathsf{t} = (\mathbf{w}_\mathsf{t}, o_\mathsf{t})$ and  $h_\mathsf{k} = (\mathbf{w}_\mathsf{k}, o_\mathsf{k})$, respectively. These normal vectors and offsets can be obtained through supervised training because the target attribute and the known attributes are pre-defined. More details of how to get these hyperplanes are shown in Appendix~\ref{subsec.gt_hyperplane}. Then, we propose the \textit{orthogonalization penalty} ($\mathcal{L}_\perp$) to tackle two problems mentioned above:
\begin{equation}
   \mathcal{L}_\perp = \mathbf{w}_\mathsf{b}^T \mathbf{w}_\mathsf{t} + \sum_{\mathsf{k} \in K} \mathbf{w}_\mathsf{b}^T \mathbf{w}_\mathsf{k}.
\end{equation}
Minimizing $\mathcal{L}_\perp$ encourages the biased attribute's hyperplane $h_\mathsf{b}$ to be orthogonalized with hyperplanes of the target attribute and known attributes. Intuitively, better orthogonalization will produce a smaller variation of traversal latent vectors' projections onto $h_\mathsf{t}$ and $h_\mathsf{k}$.

\subsection{Full Model}
\label{subsec.full_model}

Finally, we jointly minimize the \textit{total variation loss} and the \textit{orthogonality penalty} to update the hyperplane $h_\mathsf{b} = (\mathbf{w}_\mathsf{b}, o_\mathsf{b})$ (see red line in Fig.~\ref{fig.method_overview}):
\begin{equation}
   \mathcal{L} = \mathcal{L}_V+ \lambda \mathcal{L}_\perp \: ,
   \label{eq.full_model}
\end{equation}
where $\lambda$ is a coefficient of the \textit{orthogonality penalty}.

\section{Experiment}
The experiments are conducted on disentanglement datasets (Sec.~\ref{subsec.exp_disentangle_datasets}), face images (Sec.\ref{subsec.exp_celeba}), and images from other domains (\eg, cat, bedroom, \etc) (Sec.~\ref{subsec.exp_other_domain}). More details of experimental settings on each dataset will be introduced in each subsection. Additional implementation details can be seen in Appendix (Appx.)~\ref{sec.implementation_details}.

\noindent \textbf{Evaluation Metrics} As introduced in Sec.~\ref{sec.method}, we use hyperplane in the latent space to represent an attribute. For quantitative evaluation, we first choose a pair of different attributes as the ground-truth biased attribute and target attribute. Then we compute the ground-truth hyperplanes of these two attributes (more details in Appx.~\ref{subsec.gt_hyperplane}). Based on the normal vectors of hyperplanes, we design the following \textit{quantitative} evaluation metrics:

\begin{enumerate}[topsep=0pt,itemsep=-1ex,partopsep=1ex,parsep=1ex,labelindent=0pt,wide,labelwidth=!]
   \item $| \cos \langle \hat{\mathbf{w}}_\mathsf{b}, \mathbf{w}_\mathsf{b} \rangle |$ is the absolute value of cosine similarity between the predicted normal vector $\hat{\mathbf{w}}_\mathsf{b}$ and the ground-truth normal vector $\mathbf{w}_\mathsf{b}$ of biased attribute's hyperplanes. Larger $|\cos \langle \hat{\mathbf{w}}_\mathsf{b}, \mathbf{w}_\mathsf{b} \rangle|$ implies that the hyperplane prediction is closer to the ground-truth biased attribute.
   \item $| \cos \langle \hat{\mathbf{w}}_\mathsf{b}, \mathbf{w}_\mathsf{t} \rangle |$ is the absolute value of cosine similarity between the predicted normal vector $\hat{\mathbf{w}}_\mathsf{b}$ of the biased attribute's hyperplane and the ground-truth normal vector $\mathbf{w}_\mathsf{t}$ of the target attribute's hyperplane. Lower value of $\cos \langle \hat{\mathbf{w}}_\mathsf{b}, \mathbf{w}_\mathsf{t} \rangle$ means that the hyperplane prediction is more orthogonal to the target attribute hyperplane. We refer to it as ``better disentanglement \wrt the target attribute.''
   \item $\Delta \cos = | \cos \langle \hat{\mathbf{w}}_\mathsf{b}, \mathbf{w}_\mathsf{b} \rangle | - | \cos \langle \hat{\mathbf{w}}_\mathsf{b}, \mathbf{w}_\mathsf{t} \rangle |$ is difference of first two metrics. Larger values imply better results by jointly considering the first two metrics. \textbf{We use $\Delta \cos$ as the major evaluation metric} for comparing different methods because a good biased hyperplane prediction should simultaneously be closed to the ground-truth biased attribute hyperplane (\ie, large $| \cos \langle \hat{\mathbf{w}}_\mathsf{b}, \mathbf{w}_\mathsf{b} \rangle |$) and be orthogonal to the target attribute hyperplane (\ie, small $| \cos \langle \hat{\mathbf{w}}_\mathsf{b}, \mathbf{w}_\mathsf{t} \rangle |$).
   \item ``\%leading'': since each experiment contains multiple results under different settings (see \textbf{Experiment Settings} in Sec.~\ref{subsec.exp_disentangle_datasets}), we report ``\%leading'' of a method to denote the percentage of the number of settings that this method leads in terms of $\Delta \cos$.
\end{enumerate}

For the \textit{qualitative} evaluation metric, we show traversal images of different biased attribute hyperplane predictions based on the \textit{same} sampled latent code. The traversal images of the accurate biased attribute hyperplane prediction will only have variations in terms of the ground-truth biased attribute. In other words, there exists no or relatively small variations in terms of the target attribute, known attribute, or any other attributes (see examples in Fig.~\ref{fig.disentanglement_datasets}).

\begin{table}[t]
   \begin{adjustbox}{width=\columnwidth,center}
   \begin{tabular}{c|c|cccc}
   \toprule
   & method                        & $|\cos \langle \hat{\mathbf{w}}_\mathsf{b}, \mathbf{w}_\mathsf{b} \rangle|$ \(\uparrow\) & $|\cos \langle \hat{\mathbf{w}}_\mathsf{b}, \mathbf{w}_\mathsf{t} \rangle|$ \(\downarrow\) & $\Delta \cos$ \(\uparrow\) & \%leading \(\uparrow\)       \\ \midrule
   \parbox[t]{1mm}{\multirow{3}{*}{\rotatebox[origin=c]{90}{ {\scriptsize SmallNORB} }}} & VAE-based                 & 0.21\(\pm\)0.21            & 0.16\(\pm\)0.13            & 0.05$\pm$0.20                & 16.67\%     \\
   & $\mathcal{L}_H$             & \textbf{0.24\(\pm\)0.16}   & 0.26\(\pm\)0.16            & -0.02$\pm$0.24                & \textbf{31.67\%}     \\
   & \textbf{Ours}                       & 0.23\(\pm\)0.18          & \textbf{0.10\(\pm\)0.11}   & \textbf{0.12$\pm$0.21}      & \textbf{51.67\%}  \\ \midrule
   \parbox[t]{1mm}{\multirow{3}{*}{\rotatebox[origin=c]{90}{ {\small dSprites} }}} & VAE-based         & 0.11\(\pm\)0.14            & 0.13\(\pm\)0.14   & -0.01$\pm$0.16                & 22.00\%      \\
   & $\mathcal{L}_H$      & \textbf{0.23\(\pm\)0.15}   & 0.25\(\pm\)0.15            & -0.02$\pm$0.21       & \textbf{41.00\%} \\
   & \textbf{Ours}                 & 0.17\(\pm\)0.14            & \textbf{0.13\(\pm\)0.11}   & \textbf{0.05$\pm$0.18}      & \textbf{37.00\%} \\
   \bottomrule
   \end{tabular}
   \end{adjustbox}
   \caption{The mean and standard deviation results averaged over all 480 experiment settings on SmallNORB~\cite{lecun2004IEEEConf.Comput.Vis.PatternRecognit.CVPRa} and dSprites~\cite{higgins2017Int.Conf.Learn.Represent.} datasets. $\mathcal{L}_H$ denotes Hessian Penalty method. Top-2 results under \%leading metric are bolded. $\uparrow$: larger value means better result. $\downarrow$: smaller value means better result. Note that $\Delta \cos$ is the major evaluation metric that jointly considers the first two metrics. Our method achieves better performance than two baseline methods.}
   \label{tab.disentanglement_datasets_results}
   \vspace{-4mm}
\end{table}

\subsection{Experiment on Disentanglement Datasets}
\label{subsec.exp_disentangle_datasets}

\noindent \textbf{Datasets} In this experiment, we conduct experiments on two disentanglement datasets: SmallNORB~\cite{lecun2004IEEEConf.Comput.Vis.PatternRecognit.CVPRa} and dSprites~\cite{higgins2017Int.Conf.Learn.Represent.}. Both datasets contain images with a finite set of attributes, such as \textit{scale}, \textit{shape}, \etc. The numbers of attributes for two datasets are 4 and 5, respectively. We preprocess the attribute if it is not binary-valued or continuous-valued (\eg, \textit{shape}, \textit{category}), and more details of preprocessing is shown in Appx.~\ref{subsec.attr_preprocessing}.

\noindent \textbf{Generative Models}
We choose 5 VAE-based methods as the generative models: vanilla VAE~\cite{kingma2014Int.Conf.Learn.Represent.}, \(\beta\)-VAE~\cite{higgins2017Int.Conf.Learn.Represent.}, \(\beta\)-TCVAE~\cite{chen2018Adv.NeuralInf.Process.Syst.a}, DIP-VAE-I, and DIP-VAE-II~\cite{kumar2018Int.Conf.Learn.Represent.}. We use the same set of hyperparameters reported in \cite{locatelloChallengingCommonAssumptions2019} (more details in Appx.~\ref{subsec.hyperparameters}). The weights of the trained generative model is fixed and will not be updated when optimizing $h_\mathbf{b}$.

\noindent \textbf{Baseline Methods}

\noindent \textit{VAE-based}: since the aforementioned VAE-based generative models are also disentanglement methods, we use them as baseline methods. Note that these methods directly disentangle dimensions of the latent space, meaning that the normal vector of the predicted hyperplane of the biased attribute is aligned with the axis and the offset of the hyperplane is 0.

\noindent \textit{Hessian Penalty}~\cite{peebles2020Eur.Conf.Comput.Vis.ECCV} ($\mathcal{L}_H$) is another baseline method. We use the officially released implementation of the Hessian Penalty for optimizing the hyperplanes in the latent space.

\noindent More details of how to adapt baseline methods for \textit{unknown biased attribute discovery task} are described in Appx.~\ref{sec.baseline_method}.

\noindent \textbf{Experiment Settings} We create 480 experiment settings on two disentanglement datasets. Each experiment setting is a triplet of (target attribute, biased attribute, generative model) (\eg, (\textit{shape}, \textit{scale}, $\beta$-VAE)). In each setting, to make sure the target attribute classifier is biased by the chosen biased attribute, we train it on a sampled dataset with a skewed distribution between the target attribute and the biased attribute. In the example of (\textit{shape}, \textit{scale}, $\beta$-VAE) setting on dSprites dataset, the skewed training set contains more ``large heart'' images than ``small square'' images. More details
are described in Appendix~\ref{subsec.skewed_train_cls}.
For \textit{orthogonalization penalty}, we choose all remaining attributes other than biased or target attributes as the ``known attributes.''

\begin{figure}[t]
   \centering
   \includegraphics[width=\linewidth]{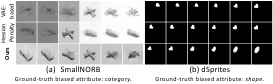}
   \caption{Qualitative comparison of traversal images on (a) SmallNORB~\cite{lecun2004IEEEConf.Comput.Vis.PatternRecognit.CVPRa} and (b) dSprites~\cite{higgins2017Int.Conf.Learn.Represent.}. The rows are traversal images based on predicted hyperplanes from different methods. The ground-truth biased attributes of (a) and (b) are \textit{category} and \textit{shape}, respectively. The target attributes of (a) and (b) are \textit{azimuth} and \textit{position x}, respectively. The traversal images from the baseline methods vary in terms of the \textit{lighting} in (a), \textit{orientation} and \textit{position x} in (b), which are different from the ground-truth biased attribute. In contrast, the traversal images from our method correctly vary in terms of the ground-truth biased attribute \textit{category} (\ie from ``car'' to ``plane'') in (a) and \textit{shape} (from ``heart'' to ``ellipse'') in (b). The quantitative results (\eg, $\Delta \cos$) of the experiment settings in this figure are in Appx.~\ref{sec.more_detailed_exp_results}.}
   \label{fig.disentanglement_datasets}
   \vspace{-6mm}
\end{figure}

\paragraph{Results}
For quantitative comparison, we report the mean and standard deviation of results over all experiment settings on each dataset. All results are summarized in Tab.~\ref{tab.disentanglement_datasets_results}. Surprisingly, the Hessian Penalty method achieves the best performance in terms of $|\cos \langle \hat{\mathbf{w}}_\mathsf{b}, \mathbf{w}_\mathsf{b} \rangle|$. However, it also achieves the worse performance in $|\cos \langle \hat{\mathbf{w}}_\mathsf{b}, \mathbf{w}_\mathsf{t} \rangle|$, implying that Hessian Penalty method learns a hyperplane that is averaged between the biased attribute and target attribute. Our method can achieve comparable results with the Hessian Penalty method in $|\cos \langle \hat{\mathbf{w}}_\mathsf{b}, \mathbf{w}_\mathsf{b} \rangle|$, and achieve the best result in $|\cos \langle \hat{\mathbf{w}}_\mathsf{b}, \mathbf{w}_\mathsf{t} \rangle|$. In terms of the major metric $\Delta \cos$ that jointly considers the first two metrics, three out of four results of the baseline methods are even negative values, meaning that their predicted hyperplanes are even closer to the target attribute hyperplanes. In contrast, our method achieves the best $\Delta \cos$ results on two datasets. In terms of \%leading, our method can also achieve the best result on SmallNORB and comparable result with Hessian Penalty on dSprites. Note that our method still achieve stabler (\ie, smaller standard deviation) $\Delta \cos$ results on dSprites. In conclusion, our proposed method can accurately discover the biased attribute and is more disentangled \wrt the target attribute. For qualitative comparison, we randomly sample an experiment setting for each dataset and generate traversal images based on the predicted hyperplanes. As shown in Fig.~\ref{fig.disentanglement_datasets}, compared with other methods, our method can accurately discover the biased attribute and the traversal images of our method do not vary in terms of the target attribute. For example, in Fig.~\ref{fig.disentanglement_datasets} (b) the traversal images of our method change by \textit{shape} (\ie, from ``heart'' to ``ellipse''), which is the ground-truth biased attribute, while traversal images of baseline methods vary over non-biased attributes such as \textit{orientation} or \textit{position x} (\ie, the object is rotating or moving horizontally).

\begin{table}[t]
   \begin{adjustbox}{width=\columnwidth,center}
   \begin{tabular}{c|cc|ccc}
   \toprule
                       & $\mathcal{L}_H$ & $\mathcal{L}_\perp$ & $|\cos \langle \hat{\mathbf{w}}_\mathsf{b}, \mathbf{w}_\mathsf{b} \rangle|$ \(\uparrow\) & $|\cos \langle \hat{\mathbf{w}}_\mathsf{b}, \mathbf{w}_\mathsf{t} \rangle|$ \(\downarrow\) & $\Delta \cos$ \(\uparrow\) \\ \midrule
   \parbox[t]{1mm}{\multirow{4}{*}{\rotatebox[origin=c]{90}{ {\small SmallNORB} }}}  &                 &               & 0.25\(\pm\)0.15             & 0.27\(\pm\)0.18             & -0.02$\pm$0.23                \\
                               &                 & \checkmark          & 0.23\(\pm\)0.18             & \textbf{0.10\(\pm\)0.11}             & \textbf{0.12$\pm$0.21}               \\
                               & \checkmark            &               & \textbf{0.27\(\pm\)0.16}             & 0.28\(\pm\)0.17             & -0.01$\pm$0.25                \\
                               & \checkmark            & \checkmark          & 0.25\(\pm\)0.17             & 0.15\(\pm\)0.13             & 0.10$\pm$0.24                \\ \midrule
   \parbox[t]{1mm}{\multirow{4}{*}{\rotatebox[origin=c]{90}{dSprites}}}   &                 &               & 0.20\(\pm\)0.13             & 0.21\(\pm\)0.13             & -0.01$\pm$0.18                \\
                               &                 & \checkmark          & 0.17\(\pm\)0.14             & \textbf{0.13\(\pm\)0.11}             & \textbf{0.05$\pm$0.18}               \\
                               & \checkmark            &               & \textbf{0.21\(\pm\)0.13}             & 0.21\(\pm\)0.13             & 0.00$\pm$0.18                \\
                               & \checkmark            & \checkmark          & \textbf{0.21\(\pm\)0.13}             & 0.19\(\pm\)0.13             & 0.01$\pm$0.18                \\
                               \bottomrule
   \end{tabular}
   \end{adjustbox}
   \caption{Ablation study on \textit{orthogonalization penalty} ($\mathcal{L}_\perp$) and Hessian Penalty~\cite{peebles2020Eur.Conf.Comput.Vis.ECCV} ($\mathcal{L}_H$). \checkmark denotes the penalty is used. Note that all rows used $\mathcal{L}_V$. We incorporate $\mathcal{L}_H$ into our method. Although adding $\mathcal{L}_H$ helps to improve $|\cos \langle \hat{\mathbf{w}}_\mathsf{b}, \mathbf{w}_\mathsf{b} \rangle|$, it seriously harms the $|\cos \langle \hat{\mathbf{w}}_\mathsf{b}, \mathbf{w}_\mathsf{t} \rangle|$. Overall, our final method (second row in each dataset) performs the best in $\Delta \cos$.}
   \label{tab.ablate_orth_hessian}
   \vspace{-3mm}
\end{table}

\noindent \textbf{Ablation Study on $\mathcal{L}_\perp$ and $\mathcal{L}_H$} We conduct ablation study on the \textit{orthogonalization penalty} ($\mathcal{L}_\perp$). The results in Tab.~\ref{tab.ablate_orth_hessian} show that $\mathcal{L}_\perp$ is helpful in decreasing $|\cos \langle \hat{\mathbf{w}}_\mathsf{b}, \mathbf{w}_\mathsf{t} \rangle|$, proving its effectiveness in being more orthogonal w.r.t. target attribute. However, $\mathcal{L}_\perp$ also decreases $|\cos \langle \hat{\mathbf{w}}_\mathsf{b}, \mathbf{w}_\mathsf{b} \rangle|$. We suggest that $\mathcal{L}_\perp$ may make the optimization problem harder due to the additional constraint. Furthermore, the good results of Hessian Penalty (denoted as $\mathcal{L}_H$) in terms of $|\cos \langle \hat{\mathbf{w}}_\mathsf{b}, \mathbf{w}_\mathsf{b} \rangle|$ motivates us to combine our method with Hessian Penalty via joint optimization. The results show that combining with $\mathcal{L}_H$ can achieve improvement in $|\cos \langle \hat{\mathbf{w}}_\mathsf{b}, \mathbf{w}_\mathsf{b} \rangle|$, as well. However, it still harms the performance in $|\cos \langle \hat{\mathbf{w}}_\mathsf{b}, \mathbf{w}_\mathsf{t} \rangle|$, suggesting that $\mathcal{L}_H$ learns an averaged hyperplane between the biased and the target attributes.
We also conduct \textbf{additional ablation studies} on the set of known attributes and the distribution of training data of generative models. Results are shown in Appx.~\ref{supp.sec.ablation_study}.

\begin{table}[t]
   \centering
   \begin{adjustbox}{width=\columnwidth,center}
       \begin{tabular}{@{}c|c|ccc@{}}
           \toprule
                           & method          & $|\cos \langle \hat{\mathbf{w}}_\mathsf{b}, \mathbf{w}_\mathsf{b} \rangle|$ \(\uparrow\)             & $|\cos \langle \hat{\mathbf{w}}_\mathsf{b}, \mathbf{w}_\mathsf{t} \rangle|$ \(\downarrow\)     & $\Delta \cos \uparrow$               \\ \midrule
           \parbox[t]{1mm}{\multirow{2}{*}{\rotatebox[origin=c]{90}{ {\scriptsize CelebA} }}} & $\mathcal{L}_H$ & 0.02$\pm$0.02         & 0.02$\pm$0.0003          & 0.0005$\pm$0.02             \\
                                   & Ours            & \textbf{0.06}$\pm$\textbf{0.01} & \textbf{0.002}$\pm$\textbf{0.001} & \textbf{0.06}$\pm$\textbf{0.01}  \\ \midrule
           \parbox[t]{1mm}{\multirow{2}{*}{\rotatebox[origin=c]{90}{ {\footnotesize FFHQ} }}}   & $\mathcal{L}_H$ & 0.05$\pm$0.01          & 0.01$\pm$0.008          & 0.03$\pm$0.004              \\
                                   & Ours            & \textbf{0.17}$\pm$\textbf{0.11} & \textbf{0.002}$\pm$\textbf{0.002} & \textbf{0.17}$\pm$\textbf{0.11} \\ \bottomrule
           \end{tabular}
   \end{adjustbox}
   \caption{Results on CelebA~\cite{liu2015IEEEInt.Conf.Comput.Vis.ICCV} and FFHQ~\cite{karras2019IEEEConf.Comput.Vis.PatternRecognit.CVPR} datasets. We omit \%leading since our method leads in all experiment settings (\ie, \%leading (Ours) = 100 \%).}
   \label{tab.celeba}
  \vspace{-5mm}
\end{table}

\begin{figure}[t]
   \centering
   \includegraphics[width=0.8\linewidth]{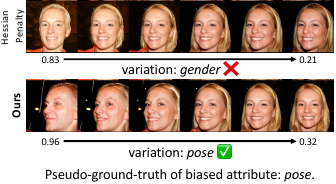}
   \caption{Qualitative comparison of the traversal images of predicted biased attribute synthesized by StyleGAN~\cite{karras2019IEEEConf.Comput.Vis.PatternRecognit.CVPR} pretrained on (a) FFHQ~\cite{karras2019IEEEConf.Comput.Vis.PatternRecognit.CVPR} dataset. The target attribute classifier is trained on FFHQ. The target attribute is \textit{gender} and the pseudo-ground-truths of biased attribute is \textit{pose}. The numbers under the images are \textit{gender} classifier's predictions on whether the \textit{gender} attribute's value is ``male.'' Our method correctly discover the \textit{pose} biased attribute.}
   \label{fig.celeba_target}
  \vspace{-5mm}
\end{figure}

\subsection{Experiment on Face Images}
\label{subsec.exp_celeba}

\noindent \textbf{Experiment Settings} We use CelebA~\cite{liu2015IEEEInt.Conf.Comput.Vis.ICCV} and FFHQ~\cite{karras2019IEEEConf.Comput.Vis.PatternRecognit.CVPR} datasets for discovering biased attributes of face images. CelebA is a dataset of face images of celebrities with 40 annotated attributes. FFHQ dataset contains 70,000 high-quality face images. We choose \textit{gender} as the target attribute to train two ResNet-18~\cite{he2016IEEEConf.Comput.Vis.PatternRecognit.CVPR} networks as the target attribute classifiers supervised by the attribute labels on two datasets, respectively. For FFHQ dataset, we use the \textit{gender} annotations from \cite{or-el2020Eur.Conf.Comput.Vis.ECCV}. Note that different from the first experiment, we do \textit{not} sample datasets with any skewed distributions because we want to discover the underlying biased attributes in the original face datasets. We choose two generative models: two StyleGAN~\cite{karras2019IEEEConf.Comput.Vis.PatternRecognit.CVPR} networks pretrained on CelebA-HQ~\cite{karras2018Int.Conf.Learn.Represent.} and FFHQ~\cite{karras2019IEEEConf.Comput.Vis.PatternRecognit.CVPR} datasets, respectively. We use the ``style'' latent space $\mathcal{W}$ of the StyleGAN, where attributes are more linearly separable than input noise latent space~\cite{karras2019IEEEConf.Comput.Vis.PatternRecognit.CVPR}. Hessian Penalty ($\mathcal{L}_H$) is used as the baseline method.
Since CelebA and FFHQ are in-the-wild datasets, we do not know the real ground-truth biased attributes. Therefore, for the quantitative evaluation, we obtain the pseudo-ground-truth of the biased attribute (see Appx.~\ref{supp.subsec.pseudo-gt_bias_attr}).
Due to the absence of real ground-truth of the biased attribute, we do not use known attributes in \textit{orthogonalization penalty} (\ie, $K = \varnothing$).

\begin{figure}[t]
   \centering
   \includegraphics[width=0.75\linewidth]{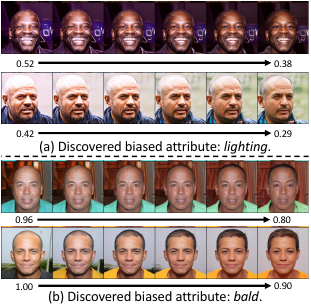}
   \caption{Discovered biased attributes by setting the set of known attributes $K$ to all considered attributes generated by StyleGAN~\cite{karras2019IEEEConf.Comput.Vis.PatternRecognit.CVPR} pretrained on FFHQ~\cite{karras2019IEEEConf.Comput.Vis.PatternRecognit.CVPR} dataset. The target attribute classifiers in (a) and (b) are trained on CelebA and FFHQ, respectively. The numbers under the images are \textit{gender} classifier's predictions on whether the \textit{gender} attribute's value is ``male.''}
   \label{fig.celeba_all}
  \vspace{-5mm}
\end{figure}

\noindent \textbf{Results} Similar to the first experiment, we run experiments under all settings of (target attribute, biased attribute, generative model). Results averaged across all settings are reported in Tab.~\ref{tab.celeba}. The results of each experiment setting are in Appx.~\ref{sec.more_detailed_exp_results}. Our method beats the Hessian Penalty method in all metrics and in each experiment setting.
The qualitative comparisons are shown in Fig.~\ref{fig.celeba_target}. The traversal images of the Hessian Penalty method vary in terms of the target attribute \textit{gender} (\ie, male to female). In contrast, our method correctly predicts the pseudo-ground-truth of the biased attribute: \textit{pose}. More examples are shown in Appx.~\ref{supp.subsec.faces_qualitative_comparison}.

\noindent \textbf{Discovering Other Biased Attributes} In this experiment, we try finding the biased attributes other than some known attributes. In the \textit{orthogonalization penalty} ($\mathcal{L}_\perp$), we let the set of the known attribute $K$ to be four attributes: \textit{age}, \textit{eyeglasses}, \textit{pose}, and \textit{smile}, whose hyperplanes are provided in \cite{shen2020IEEEConf.Comput.Vis.PatternRecognit.CVPR}. Results in Fig.~\ref{fig.celeba_all} show that our method can successfully discover other biased attributes such as \textit{lighting} and \textit{bald}, proving the effectiveness of the known attributes in $\mathcal{L}_\perp$. One may regard the variations that do not switch 0.5-threshold (\eg, 1.00 to 0.90 in Fig.~\ref{fig.celeba_all}) are not meaningful. We believe such variation is still valuable and include a discussion in Appx.~\ref{supp.subsec.small_variation}. We also conduct a user study (see Appx.~\ref{supp.sec.user_study}) to invite subjects to name the attributes from more traversal images, whose results prove that our method finds biased attributes difficult for the Hessian Penalty method.

\begin{figure}[t]
   \centering
   \includegraphics[width=0.85\linewidth]{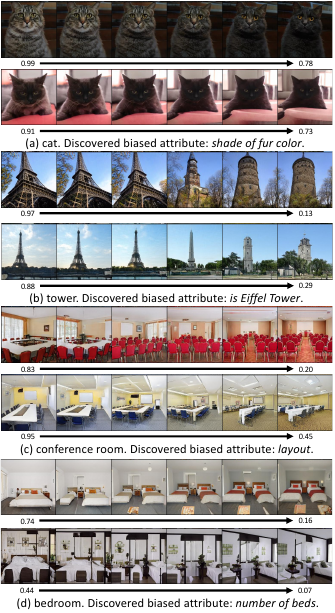}
   \caption{Discovered biased attribute of classifiers for classifying \textit{cat}, \textit{tower}, \textit{conference room}, \textit{bedroom} images. Numbers below images are predicted probability by the classifier.}
   \label{fig.other_domains}
\end{figure}

\subsection{Experiment on Images from Other Domains}
\label{subsec.exp_other_domain}

Finally, we apply our method on images from other domains, including object (\eg, \textit{cat}) and scenes (\eg, \textit{bedroom}) categories in LSUN dataset~\cite{yu2016ArXiv150603365Cs}. StyleGAN and StyleGAN2 pretrained on the images from each category are used as generative models and weights are obtained from~\cite{shen2020}. Since each generator is trained on only one category of images (\eg, the cat generator is only trained on cat images), we only use $\mathcal{L}_V$ and do not use $\mathcal{L}_\perp$ because the target attribute value (\ie, object or scene category) is fixed for each generator. We choose ResNet-18~\cite{he2016IEEEConf.Comput.Vis.PatternRecognit.CVPR} pretrained on ImageNet as the object classifier and ResNet-18 pretrained on Places365 as the scene classifier. We show some biased attributed discovered by our method in Fig.~\ref{fig.other_domains}. Our method successfully discovers biased attributes such as \textit{shade of fur color}, \textit{is Eiffel Tower}, \textit{layout}, and \textit{number of beds} in cat, tower, conference room, and bedroom classifiers, respectively, which could be hard for human to speculate in advance. We also conduct a user study (see Appx.~\ref{supp.sec.user_study}) on letting subjects name the biased attributes from more traversal images, which verifies our method discovers biased attributes that are difficult for Hessian Penalty. This proves the generalizability of our method for discovering biased attributes in various image domains.

\section{Conclusion}
In this work, we propose a new problem for finding the unknown biased attribute of a classifier without presumptions or labels. To tackle this new problem, a novel method is proposed for this task by optimizing \textit{total variation loss} and \textit{orthogonalization penalty}. The comprehensive experiments prove that our method is effective and can discover biased attributes in multiple domains. In the appendix, we discuss the limitations, future directions, and the related methods and areas that can be benefited from this new \textit{unknown biased attribute discovery task}.

\paragraph{Acknowledgements.}
This work has been partially supported by the National Science Foundation (NSF) under Grants 1764415, 1813709, 1909912, and 1934962. The article solely reflects the opinions and conclusions of its authors but not the funding agents.

\clearpage
{\small
\bibliographystyle{ieee_fullname}
\bibliography{main_bib}
}

\section*{Appendix}
\appendix
\section{Implementation Details}
\label{sec.implementation_details}

Here we introduce the implementation details. All experiments are implemented in PyTorch~\cite{paszke2019Adv.NeuralInf.Process.Syst.}.

\subsection{Hyperparameters}
\label{subsec.hyperparameters}
The following hyperparameters in this paragraph are used across all experiments on all datasets. We use Adam~\cite{kingma2015Int.Conf.Learn.Represent.} to optimize the biased attribute hyperplane $h_\mathsf{b}$, where $\beta_1$, and $\beta_2$ are $0.9$, and $0.999$, respectively. The number of iterations for both Hessian Penalty~\cite{peebles2020Eur.Conf.Comput.Vis.ECCV} and our method is $1000$.

In the experiments on disentanglement datasets (Sec.~\ref{subsec.exp_disentangle_datasets}), we set the batch size to $64$. The learning rate is $10^{-3}$. The coefficient of \textit{orthogonalization penalty} (see $\lambda$ in Eq.~\ref{eq.full_model}) is $10$. In terms of traversal images, we set the number of traversal steps $N$ (see $N$ in Eq.~\ref{eq.total_variation_loss}) to 20, where the step sizes $\{\alpha_i \mid i = 1 \ldots N\}$ (see $\alpha_i$ in Eq.~\ref{eq.total_variation_loss}) are numbers evenly spaced in the interval $[-2, 2]$.

In the experiments on face images (Sec.~\ref{subsec.exp_celeba}), the learning rate is $10^{-1}$. The coefficient of \textit{orthogonalization penalty} is $100$. In the experiments on images from other domains (Sec.~\ref{subsec.exp_other_domain}), the learning rate is $10^{-3}$. In both previous experiments, we set the batch size to $1$ and set the number of traversal steps $N$ (see $N$ in Eq.~\ref{eq.total_variation_loss}) to 6, where the step sizes $\{\alpha_i \mid i = 1 \ldots N\}$ (see $\alpha_i$ in Eq.~\ref{eq.total_variation_loss}) are numbers evenly spaced in the interval $[-3, 3]$.

\subsection{Projecting Latent Code to Hyperplane}
\label{subsec.project_code_to_plane}
We use the offset $o_b$ for projecting randomly sampled latent code $\mathbf{z}$ to the hyperplane $h_\mathsf{b}$. To explain that $o_b$ is jointly optimized with $\mathbf{w}_\mathsf{b}$ when minimizing the \textit{total variation loss} $\mathcal{L}_V$, we describe the complete algorithm for computing \textit{total variation loss} $\mathcal{L}_V$.

We define the biased attribute's hyperplane $h_\mathsf{b}$ as $\mathbf{w}_\mathsf{b}^T \mathbf{x} + o_\mathsf{b} = 0$. Therefore, the projected sampled latent code $\mathbf{z}_\text{proj}$ can be computed by:
\begin{equation}
    \mathbf{z}_\text{proj} = \mathbf{z} - \frac{\mathbf{w}_\mathsf{b}^T \mathbf{z} + o_\mathsf{b}}{||\mathbf{w}_\mathsf{b}\||^2} \mathbf{w}_\mathsf{b}.
\end{equation}
Then, the projected latent code $\mathbf{z}_\text{proj}$ is used to sample traversal latent codes $ \{\mathbf{z}_\text{proj} + \alpha_i \frac{\mathbf{w}_\mathbf{b}}{||\mathbf{w}_\mathbf{b}||} \mid i = 1 \ldots N \} $, which are fed to generative model $\mathit{G}$ for synthesizing traversal images:
\begin{equation}
    \mathbf{I}(\mathsf{b} = b_i) = \mathit{G} (\mathbf{z}_\text{proj} + \alpha_i \frac{ \mathbf{w}_b }{ ||\mathbf{w}_b||}).
\end{equation}
Next, the traversal images $\{\mathbf{I}(\mathsf{b} = b_i) \mid i = 1 \ldots N\}$ are fed to the classifier for predicting the target attributes $\{ \mathit{P}(\hat{\mathsf{t}} \mid \mathbf{I}(\mathsf{b} = b_i) ) \mid 1 \ldots N \}$. Finally, the \textit{total variation loss} can be computed by:
\begin{align}
    \begin{split}
        \mathcal{L}_V(\mathbf{w}_\mathsf{b}, o_\mathsf{b}) = \\
         -\log \frac{1}{N-1} \sum_{i=1}^{N-1} | & \mathit{P}(\hat{\mathsf{t}} \mid \mathit{G}(\mathbf{z}_\text{proj}+ \alpha_{i+1} \frac{\mathbf{w}_\mathsf{b}}{||\mathbf{w}_\mathsf{b}||})) \\ & - \mathit{P}(\hat{\mathsf{t}} \mid \mathit{G}(\mathbf{z}_\text{proj} + \alpha_i \frac{\mathbf{w}_\mathsf{b}}{||\mathbf{w}_\mathsf{b}||}))|.
    \end{split}
    \label{eq.tv_loss}
\end{align}
Note that the \textit{total variation loss} $\mathcal{L}_V$ is a function of both $\mathbf{w}_\mathsf{b}$ and $o_\mathsf{b}$ because $\mathbf{z}_\text{proj}$ is computed based on both of them. Therefore, we can optimize both $\mathbf{w}_\mathsf{b}$ and $o_\mathsf{b}$ when minimizing the \textit{total variation loss} $\mathcal{L}_V$.

\begin{algorithm}
    \DontPrintSemicolon
      \KwInput{$\{\mathbf{I}, \mathbf{a}\}$: set of pairs of image $\mathbf{I} \in \mathbb{R}^{H \times W \times 3}$ and attribute label $\mathbf{a} \in \mathbb{R}^{J}$ ($J$: number of attributes); $E: \mathbb{R}^{H \times W \times 3} \rightarrow \mathbb{R}^{d}$: image encoder of VAE-based model.}
      \KwOutput{$Q \in \mathbb{R}^{d \times J}$: normal vectors of attribute hyperplane; $\mathbf{o} \in \mathbb{R}^J$: offsets of attribute hyperplane}
      $\{\mathbf{z}\} \coloneqq E(\{\mathbf{I}\})$ \tcc{Encode all images into the latent space.}
      randomly initialize a matrix $W \in \mathbb{R}^{d \times J}$ \\
      randomly initialize a vector $\mathbf{o} \in \mathbb{R}^J$ \\
      \For{\text{each iteration}}
      {
        $Q \coloneqq \text{QR-decomposition}(W)$ \\
        $\mathbf{p} \coloneqq \text{sigmoid}(Q^T\mathbf{z} + \mathbf{o})$ \\
        $l \coloneqq \text{BCE}(\mathbf{p}, \mathbf{a}) \: \text{or MSE}(\mathbf{p}, \mathbf{a})$ \tcc{use binary cross-entropy (BCE) loss for binary attribute and mean squared error for continuous valued attribute}
        $W, \mathbf{o} \coloneqq \text{Adam}(l)$ \tcc{update with Adam optimizer}
      }
    \caption{Compute Ground-truth Hyperplanes}
    \label{alg.gt_hyperplane}
\end{algorithm}

\subsection{Ground-truth Hyperplanes on Disentanglement Datasets}
\label{subsec.gt_hyperplane}
Here we describe how to compute the ground-truth hyperplane for all attributes in the dataset. We do not follow previous works~\cite{karras2019IEEEConf.Comput.Vis.PatternRecognit.CVPR,shen2020IEEEConf.Comput.Vis.PatternRecognit.CVPR,balakrishnan2020Eur.Conf.Comput.Vis.ECCV} to compute the hyperplane via SVM~\cite{cortes1995MachLearn} or logistic regression~\cite{nelder1972J.R.Stat.Soc.Ser.Gen.} because we observe a strong correlation of variation from different attributes even with the orthogonalization trick introduced by \citet{balakrishnan2020Eur.Conf.Comput.Vis.ECCV}. We suspect that the correlation exists because the hyperplanes are computed individually. Therefore, we propose a new method to optimize all hyperplanes jointly. The algorithm is described in Algorithm~\ref{alg.gt_hyperplane}.

Suppose the dataset has $J$ attributes (\ie, $J = 4, J=5$ for SmallNORB~\cite{lecun2004IEEEConf.Comput.Vis.PatternRecognit.CVPRa} and dSprites~\cite{higgins2017Int.Conf.Learn.Represent.}, respectively) and the dimension of VAE-based model's latent space is $d$. We are also given the dataset's images $\{\mathbf{I}\}$ and corresponding attributes labels $\{\mathbf{a} \in \mathbb{R}^{J}\}$. Then we encode the images $\{\mathbf{I}\}$ to latent codes $\{\mathbf{z}\}$ by the pre-trained encoder of the VAE-based model. We will use latent codes $\{\mathbf{z}\}$ paired with the attribute labels $\{\mathbf{a} \in \mathbb{R}^{J}\}$ as training data to compute the ground-truth hyperplanes.

After obtaining the training data, we initialize a matrix $W \in \mathbb{R}^{d \times J}$, where the $j$-th column in $W$ represents the normal vector of $j$-th attribute's hyperplane. Similarly, we initialize a vector $\mathbf{o} \in \mathbb{R}^J$, where $j$-th value in $\mathbf{o}$ represents the offset of $j$-th attribute's hyperplane. Then, we perform QR-decomposition on $W$ to obtain the orthogonal matrix $Q = \text{QR-decomposition}(W)$. Next, $Q$ and $\mathbf{o}$ are used to classify the latent codes $\{\mathbf{z}\}$. Supervised by the attribute labels $\{\mathbf{a}\}$, we optimize $W$ and $\mathbf{o}$ iteratively via Adam optimizer. After the optimization, $Q$ and $\mathbf{o}$ are the ground-truth normal vectors and offsets of hyperplanes for all attributes in the given dataset. Since we use five different VAE-based models to compute latent codes $\{\mathbf{z}\}$, the ground-truth hyperplane of the same attribute is different for different VAE-based models.

The computed ground-truth hyperplanes by the method mentioned above are used for evaluation. However, we cannot use them in the \textit{orthogonalization penalty}. The reason is that the orthogonal matrix $Q$ ensures the orthogonalization among the hyperplanes, which cannot be realized in the real-world setting where the biased attribute is \textit{unknown} so that we cannot let the hyperplanes of the target attribute and known attributes be orthogonal with the \textit{unknown} biased attribute. Therefore, the normal vectors for the \textit{orthogonalization penalty} are computed in a different way. Suppose the $b$-th attribute is selected as the ground-truth biased attribute under an experimental setting (recall that an experimental setting is a triplet of (target attribute, biased attribute, generative model)). Then, we remove the $b$-th column in the optimized $W$ to obtain a new matrix $W' \in \mathbb{R}^{J-1}$. Next, we apply the QR-decomposition to obtain $Q'$ (\ie, $Q' = \text{QR-docomposition}(W')$), so that column vectors in $Q' \in \mathbb{R}^{J-1}$ are used as normal vectors of hyperplanes for target attribute and known attributes in the \textit{orthogonalization penalty}.

\subsection{Pseudo-ground-truth of Biased Attribute on Face Images}
\label{supp.subsec.pseudo-gt_bias_attr}
As described in the main paper, since CelebA~\cite{liu2015IEEEInt.Conf.Comput.Vis.ICCV} and FFHQ~\cite{karras2019IEEEConf.Comput.Vis.PatternRecognit.CVPR} are in-the-wild datasets, we do not know the real ground-truth biased attributes. Therefore, for the quantitative evaluation, we obtain the pseudo-ground-truth of the biased attribute as follows: First, we assume a larger set of attributes as the potential biased attribute by adding \textit{age}, \textit{smile}, \textit{eyeglasses}, and \textit{pose} attributes into consideration. Then, we obtain the ground-truth hyperplanes of all five attributes (four attributes mentioned before plus the \textit{gender} attribute) in the latent space of StyleGAN from \citet{shen2020IEEEConf.Comput.Vis.PatternRecognit.CVPR}. Next, for each pair of a target attribute and a possible biased attribute (pairs of identical attributes are excluded), we generate traversal images based on the biased attribute, test them on a target attribute classifier, and record the total variation (TV). We pick the attribute that produces the largest total variation (TV) as the pseudo-ground-truth for the target attribute.

\subsection{Generative Models}
\paragraph{Experiments on Disentanglement Datasets}
We use the same set of hyperparameters with disentanglement-lib~\cite{locatelloChallengingCommonAssumptions2019} to train VAE-based models and we use Disentanglement-PyTorch~\cite{abdi2019Adv.NeuralInf.Process.Syst.Workshop} as the code base for training. The dimension of latent space of all VAE-based generative models (vanilla VAE~\cite{kingma2014Int.Conf.Learn.Represent.}, \(\beta\)-VAE~\cite{higgins2017Int.Conf.Learn.Represent.}, DIP-VAE-I, and DIP-VAE-II~\cite{kumar2018Int.Conf.Learn.Represent.}) is 10. The image size is $64 \times 64$. The optimizer for training VAE-based models is Adam ($\beta_1=0.9, \beta_2=0.999$) and the learning rate is $10^{-4}$. The training steps is $3 \times 10^5$.

\paragraph{Experiments on Face Images}
We use two generative models: two StyleGANs\cite{karras2019IEEEConf.Comput.Vis.PatternRecognit.CVPR} models trained on CelebA-HQ~\cite{karras2018Int.Conf.Learn.Represent.} and FFHQ~\cite{karras2019IEEEConf.Comput.Vis.PatternRecognit.CVPR}, respectively. The weights are obtained from the officially released code from \citet{shen2020IEEEConf.Comput.Vis.PatternRecognit.CVPR}. The dimension of latent space of models mentioned above is $512$. The latent space of all models is the $\mathcal{W}$-space. The synthesized image size is $1024 \times 1024$ and we resize it to $64 \times 64$ before feed them to the classifier.

\paragraph{Experiments on Images from Other Domains}
StyleGAN~\cite{karras2019IEEEConf.Comput.Vis.PatternRecognit.CVPR} and StyleGAN2~\cite{karras2020IEEEConf.Comput.Vis.PatternRecognit.CVPR} trained on the images from each category in LSUN~\cite{yu2016ArXiv150603365Cs} are used as generative models and weights are obtained from \citet{shen2020}. The size of the synthesized images is $256 \times 256$. We use $\mathcal{Z}$-space as the latent space of the generated model, where the dimension is $512$.

\subsection{Baseline Methods}
\label{sec.baseline_method}

\paragraph{Adaptation of Baseline Methods}
We choose unsupervised disentanglement methods as baselines. Here, we describe how to adapt them for \textit{unknown biased attribute discovery task}. First, a trained unsupervised disentanglement method will predict a set of hyperplanes in the latent space. Then, we remove the hyperplane whose normal vector has the largest absolute value of cosine similarity with the normal vector of ground-truth target attribute hyperplane because it can be regarded as the predicted hyperplane for the target attribute. Then, we use each of the remaining hyperplanes to generate batches of traversal images, feed them to the classifier, and record the average total variation (TV, which will be introduced in Sec.~\ref{sec.total_variation}) over each batch. The hyperplane with the largest average TV is selected as the predicted biased attribute hyperplane.

\paragraph{Results of VAE-based Method}
We create 480 different experiment settings in the experiments on disentanglement datasets, where each setting is a triplet of (target attribute, biased attribute, generative model). To help readers better understand how to compute the results for the VAE-based method, we use the following example to illustrate. Under the setting (\textit{shape}, \textit{scale}, $\beta$-VAE), we use the predicted axis-aligned hyperplane of $\beta$-VAE as the prediction for the VAE-based method. In other words, under each setting, the result of the VAE-based method depends on the generative model used in the experiment setting.

\subsection{Attributes Preprocessing}
\label{subsec.attr_preprocessing}
We preprocess the attributes' values in disentanglement datasets if they are not binary-valued or continuous-valued. Here we introduce the details of attributes preprocessing.

For \textit{shape} and \textit{category} attributes in the disentanglement datasets, we choose a subset of shapes or categories as the positive class and others belong to the negative class. Therefore, we convert the \textit{shape} and \textit{category} attributes to binary-valued attributes. Concretely, for \textit{category} attribute in SmallNORB~\cite{lecun2004IEEEConf.Comput.Vis.PatternRecognit.CVPRa}, we choose ``square'' and ``ellipse'' as the positive class and ``heart'' as the negative class. For \textit{shape} attribute in dSprites~\cite{higgins2017Int.Conf.Learn.Represent.}, we choose ``four-legged animals'' and ``human figures'' as the positive class and ``airplanes,'' ``trucks,'' and ``cars'' as the negative class.

\begin{table*}[t]
   \centering
   \begin{tabular}{c|c|ccccc}
   \toprule
   dataset & method                        & $|\cos \langle \hat{\mathbf{w}}_\mathsf{b}, \mathbf{w}_\mathsf{b} \rangle|$ \(\uparrow\) & $|\cos \langle \hat{\mathbf{w}}_\mathsf{b}, \mathbf{w}_\mathsf{t} \rangle|$ \(\downarrow\) & $\Delta \cos$ \(\uparrow\) & \%leading \(\uparrow\)  & TV      \\ \midrule
   \multirow{3}{*}{SmallNORB} & VAE-based                 & 0.21\(\pm\)0.21            & 0.16\(\pm\)0.13            & 0.05$\pm$0.20                & 16.67\%  & 0.15$\pm$0.07        \\
   & Hessian Penalty             & \textbf{0.24\(\pm\)0.16}   & 0.26\(\pm\)0.16            & -0.02$\pm$0.24                & \textbf{31.67\%}  &  0.14$\pm$0.05     \\
   & \textbf{Ours}                       & 0.23\(\pm\)0.18          & \textbf{0.10\(\pm\)0.11}   & \textbf{0.12$\pm$0.21}      & \textbf{51.67\%} &  0.13$\pm$0.05 \\ \midrule
   \multirow{3}{*}{dSprites} & VAE-based         & 0.11\(\pm\)0.14            & 0.13\(\pm\)0.14   & -0.01$\pm$0.16                & 22.00\%     &  0.09$\pm$0.05   \\
   & Hessian Penalty      & \textbf{0.23\(\pm\)0.15}   & 0.25\(\pm\)0.15            & -0.02$\pm$0.21       & \textbf{41.00\%} & 0.09$\pm$0.04 \\
   & \textbf{Ours}                 & 0.17\(\pm\)0.14            & \textbf{0.13\(\pm\)0.11}   & \textbf{0.05$\pm$0.18}      & \textbf{37.00\%} & 0.07$\pm$0.04 \\
   \bottomrule
   \end{tabular}
   \caption{Tab.~\ref{tab.disentanglement_datasets_results} in the main paper with Total Variation (TV) results. The TV metric is introduced in Sec.~\ref{sec.total_variation}.  We do not bold the TV results because the baseline methods achieve larger $|\cos \langle \hat{\mathbf{w}}_\mathsf{b}, \mathbf{w}_\mathsf{t} \rangle|$ (see explanations in Sec.~\ref{sec.total_variation}). The table shows mean and standard deviation results averaged over all 480 experiment settings on SmallNORB~\cite{lecun2004IEEEConf.Comput.Vis.PatternRecognit.CVPRa} and dSprites~\cite{higgins2017Int.Conf.Learn.Represent.} datasets. Top-2 results under \%leading metric are bolded. $\uparrow$: larger value means better result. $\downarrow$: smaller value means better result. Note that $\Delta \cos$ is the major evaluation metric that jointly considers the first two metrics. Our method achieves better performance than two baseline methods.}
   \label{supp.tab.disentanglement_datasets_results}
\end{table*}

\begin{table*}[t]
    \centering
    \begin{tabular}{c|cc|cccc}
        \toprule
                            & $\mathcal{L}_H$ & $\mathcal{L}_\perp$ & $|\cos \langle \hat{\mathbf{w}}_\mathsf{b}, \mathbf{w}_\mathsf{b} \rangle|$ \(\uparrow\) & $|\cos \langle \hat{\mathbf{w}}_\mathsf{b}, \mathbf{w}_\mathsf{t} \rangle|$ \(\downarrow\) & $\Delta \cos$ \(\uparrow\) & TV \\ \midrule
        \parbox[t]{1mm}{\multirow{4}{*}{\rotatebox[origin=c]{90}{ {\small SmallNORB} }}}  &                 &               & 0.25\(\pm\)0.15             & 0.27\(\pm\)0.18             & -0.02$\pm$0.23              & 0.15$\pm$0.05 \\
                                    &                 & \checkmark          & 0.23\(\pm\)0.18             & \textbf{0.10\(\pm\)0.11}             & \textbf{0.12$\pm$0.21}              &  0.13$\pm$0.05 \\
                                    & \checkmark            &               & \textbf{0.27\(\pm\)0.16}             & 0.28\(\pm\)0.17             & -0.01$\pm$0.25    &  0.10$\pm$0.03          \\
                                    & \checkmark            & \checkmark          & 0.25\(\pm\)0.17             & 0.15\(\pm\)0.13             & 0.10$\pm$0.24   &     0.10$\pm$0.03        \\ \midrule
        \parbox[t]{1mm}{\multirow{4}{*}{\rotatebox[origin=c]{90}{dSprites}}}   &                 &               & 0.20\(\pm\)0.13             & 0.21\(\pm\)0.13             & -0.01$\pm$0.18       &  0.07$\pm$0.04       \\
                                    &                 & \checkmark          & 0.17\(\pm\)0.14             & \textbf{0.13\(\pm\)0.11}             & \textbf{0.05$\pm$0.18}         &  0.07$\pm$0.04    \\
                                    & \checkmark            &               & \textbf{0.21\(\pm\)0.13}             & 0.21\(\pm\)0.13             & 0.00$\pm$0.18  &    0.06$\pm$0.04          \\
                                    & \checkmark            & \checkmark          & \textbf{0.21\(\pm\)0.13}             & 0.19\(\pm\)0.13             & 0.01$\pm$0.18        &  0.06$\pm$0.04      \\
                                    \bottomrule
        \end{tabular}
    \caption{Tab.~\ref{tab.ablate_orth_hessian} in the main paper with Total Variation (TV) results. The TV metric is introduced in Sec.~\ref{sec.total_variation}. The table shows the ablation study on \textit{orthogonalization penalty} ($\mathcal{L}_\perp$) and Hessian Penalty~\cite{peebles2020Eur.Conf.Comput.Vis.ECCV} ($\mathcal{L}_H$). \checkmark denotes the penalty is used. Note that all rows used $\mathcal{L}_V$. We incorporate $\mathcal{L}_H$ into our method. Although adding $\mathcal{L}_H$ helps to improve $|\cos \langle \hat{\mathbf{w}}_\mathsf{b}, \mathbf{w}_\mathsf{b} \rangle|$, it seriously harms the $|\cos \langle \hat{\mathbf{w}}_\mathsf{b}, \mathbf{w}_\mathsf{t} \rangle|$. Overall, our final method (second row in each dataset) performs the best in $\Delta \cos$.}
    \label{supp.tab.ablate_orth_hessian}
\end{table*}

\begin{table*}[t]
    \centering
    \begin{tabular}{@{}c|c|cccc@{}}
        \toprule
                        & method          & $|\cos \langle \hat{\mathbf{w}}_\mathsf{b}, \mathbf{w}_\mathsf{b} \rangle|$ \(\uparrow\)             & $|\cos \langle \hat{\mathbf{w}}_\mathsf{b}, \mathbf{w}_\mathsf{t} \rangle|$ \(\downarrow\)     & $\Delta \cos \uparrow$       & TV        \\ \midrule
        \parbox[t]{1mm}{\multirow{2}{*}{\rotatebox[origin=c]{90}{ {\scriptsize CelebA} }}} & $\mathcal{L}_H$ & 0.02$\pm$0.02         & 0.02$\pm$0.0003          & 0.0005$\pm$0.02    &   0.02$\pm$0.01       \\
                                & Ours            & \textbf{0.06}$\pm$\textbf{0.01} & \textbf{0.002}$\pm$\textbf{0.001} & \textbf{0.06}$\pm$\textbf{0.01} &   \textbf{0.11$\pm$0.008} \\ \midrule
        \parbox[t]{1mm}{\multirow{2}{*}{\rotatebox[origin=c]{90}{ {\footnotesize FFHQ} }}}   & $\mathcal{L}_H$ & 0.05$\pm$0.01          & 0.01$\pm$0.008          & 0.03$\pm$0.004          &  0.05$\pm$0.001    \\
                                & Ours            & \textbf{0.17}$\pm$\textbf{0.11} & \textbf{0.002}$\pm$\textbf{0.002} & \textbf{0.17}$\pm$\textbf{0.11} &   \textbf{0.11$\pm$0.001} \\ \bottomrule
    \end{tabular}
    \caption{Tab.~\ref{tab.celeba} in the main paper with Total Variation (TV) results. The table shows the results on CelebA~\cite{liu2015IEEEInt.Conf.Comput.Vis.ICCV} and FFHQ~\cite{karras2019IEEEConf.Comput.Vis.PatternRecognit.CVPR} datasets. We omit \%leading since our method leads in all experiment settings (\ie, \%leading (Ours) = 100 \%). We bold the TV results because our method also achieves smaller $|\cos \langle \hat{\mathbf{w}}_\mathsf{b}, \mathbf{w}_\mathsf{t} \rangle|$ (see explanation in Sec.~\ref{sec.total_variation}).}
    \label{supp.tab.celeba}
 \end{table*}

\subsection{Training Biased Classifiers on Disentanglement Datasets}
\label{subsec.skewed_train_cls}
In order to ensure that the classifier is biased to the chosen biased attribute (\ie, ground-truth biased attribute), following the method in \cite{wang2020IEEEConf.Comput.Vis.PatternRecognit.CVPRe}, we sample the disentanglement dataset with skewed distribution to train the classifier. Formally, we denote the biased attribute as $\mathsf{b}$ and the target attribute as $\mathsf{t}$. First, for the sampling purpose, we transform the target attribute and the biased attribute to the binary-valued attributes if they are continuous-valued attributes. We achieve this by considering the values less than the medium value as the positive class and the values greater or equal than the medium value as the negative class. Note that such binary-valued attributes are only used for sampling and will not be used for training. Second, we uniformly sample the binary value of the target attribute (\ie, $\mathsf{t} = 0$ or $\mathsf{t} = 1$). Next, we sample the value of the biased attribute based on the following skewed conditional distribution:
\begin{align*}
    P(\mathsf{b}=0 \mid \mathsf{t}=1) &= S\\
    P(\mathsf{b}=1 \mid \mathsf{t}=1) &= 1 - S\\
    P(\mathsf{b}=0 \mid \mathsf{t}=0) &= 1 - S\\
    P(\mathsf{b}=1 \mid \mathsf{t}=0) &= S,
\end{align*}
where $S$ is the ``skewness'' of the conditional distribution for sampling the biased attribute. We set $S = 0.9$ in all experiments on the disentanglement dataset. The ablation study on ``skewness'' is shown in Sec.~\ref{subsec.ablate_skewness}. After sampling the values for the biased attribute and the target attribute, we use them to uniformly sample the data with the sampled values in terms of the biased attribute and the target attribute from the dataset.

\section{Evaluation Metric - Total Variation (TV)}
\label{sec.total_variation}

\subsection{Definition of TV}
We also report the results with another evaluation metric -- Total Variation (TV), which is formally defined by:
\begin{align}
    \begin{split}
        \text{TV}(\mathbf{w}_\mathsf{b}, o_\mathsf{b}) = \frac{1}{N-1} \sum_{i=1}^{N-1} | & \mathit{P}(\hat{\mathsf{t}} \mid \mathit{G}(\mathbf{z}_\text{proj}+ \alpha_{i+1} \frac{\mathbf{w}_\mathsf{b}}{||\mathbf{w}_\mathsf{b}||})) \\ & - \mathit{P}(\hat{\mathsf{t}} \mid \mathit{G}(\mathbf{z}_\text{proj} + \alpha_i \frac{\mathbf{w}_\mathsf{b}}{||\mathbf{w}_\mathsf{b}||}))|.
    \end{split}
    \label{eq.tv_metric}
\end{align}
Compared with the definition of \textit{total variation loss} (Eq.~\ref{eq.tv_loss}), TV removes the $- \log$. Intuitively, TV captures the averaged absolute difference of classifier's predictions between each pair of consecutive steps. Therefore, larger TV values indicate larger variations of target attribute classifier's predictions on the traversal images. We add TV metric results of Tab.~\ref{tab.disentanglement_datasets_results}, \ref{tab.ablate_orth_hessian}, \ref{tab.celeba} in Tab.~\ref{supp.tab.disentanglement_datasets_results}, \ref{supp.tab.ablate_orth_hessian}, \ref{supp.tab.celeba}, respectively. We also report the TV results of the ground-truth biased attribute and the target attribute hyperplanes on disentanglement datasets and face image datasets in Tab.~\ref{tab.gt_tv_disentanglement_datasets} and Tab.~\ref{tab.gt_tv_face_datasets}, respectively. To compute the TV of the target attribute hyperplane, we replace $\mathbf{w}_\mathsf{b}$ with $\mathbf{w}_\mathsf{t}$ in Eq.~\ref{eq.tv_metric}. Note that the TV values of predicted biased attribute from all methods are larger than the TV values of ground-truth biased attribute and ground-truth target attribute. We believe the reason is that all methods' predicted biased attribute hyperplanes are not perfectly orthogonal \wrt the ground-truth target attribute. Hence, both biased attribute and target attribute will vary in the traversal images, leading to larger classifier prediction variations than the ground-truth hyperplanes that only have single-attribute variations.

\subsection{TV as an Unfairness Metric}
Note that two cases of biased attribute prediction will cause large TV values:
\begin{enumerate}
    \item the prediction is close to ground-truth of the biased attribute;
    \item the prediction is close to the target attribute (\ie, trivial solution explained in Sec.~\ref{subsec.orth_penalty}).
\end{enumerate}
Therefore, \textbf{we regard TV as an \textit{unfairness metric} only when the method has smaller $| \cos \langle \hat{\mathbf{w}}_\mathsf{b}, \mathbf{w}_\mathsf{t} \rangle |$} (\ie, nontrivial solution). In other words, larger TV value does not mean better result if the method also has larger $| \cos \langle \hat{\mathbf{w}}_\mathsf{b}, \mathbf{w}_\mathsf{t} \rangle |$. For example, in Tab.~\ref{supp.tab.disentanglement_datasets_results}, although baseline methods have larger TV results, they also have larger $| \cos \langle \hat{\mathbf{w}}_\mathsf{b}, \mathbf{w}_\mathsf{t} \rangle |$. Therefore, baseline methods' biased attribute predictions are not more unfair than our method, but rather closer to the trivial solution. However, comparing the TV result with Hessian Penalty on face image datasets in Tab.~\ref{supp.tab.celeba}, our method achieves larger TV and smaller $| \cos \langle \hat{\mathbf{w}}_\mathsf{b}, \mathbf{w}_\mathsf{t} \rangle |$ simultaneously. Therefore, on face image datasets, our method not only accurately predicts the biased attribute with larger unfairness results (\ie, TV), but also avoids the trivial solution.

\begin{table}[t]
    \centering
    \begin{tabular}{@{}c|cc@{}}
    \toprule
    dataset   & GT Biased TV & GT Target TV \\ \midrule
    SmallNORB & 0.06$\pm$0.06   & 0.05$\pm$0.04   \\
    dSprites  & 0.03$\pm$0.04   & 0.04$\pm$0.03   \\ \bottomrule
    \end{tabular}
    \caption{Total variation (TV) of the ground-truth biased attribute (GT Biased TV) and the ground-truth target attribute (GT Target TV) on SmallNORB and dSprites datasets.}
    \label{tab.gt_tv_disentanglement_datasets}
\end{table}

\begin{table}[t]
    \centering
    \begin{tabular}{@{}c|c|cc@{}}
    \toprule
    classifier              & StyleGAN  & PGT Biased TV & GT Target TV \\ \midrule
    \multirow{2}{*}{CelebA} & CelebA-HQ & 0.07       & 0.05      \\
                            & FFHQ      & 0.07       & 0.04      \\ \midrule
    \multirow{2}{*}{FFHQ}   & CelebA-HQ & 0.10       & 0.04      \\
                            & FFHQ      & 0.06       & 0.11      \\ \bottomrule
    \end{tabular}
    \caption{Total variation (TV) of the pseudo-ground-truth of biased attribute (PGT Biased TV) and ground-truth target attribute (GT Target TV) on face image datasets. The first two columns denote the training datasets of the target attribute classifier and StyleGAN, respectively.}
    \label{tab.gt_tv_face_datasets}
\end{table}

\begin{table}[t]
    \centering
    \begin{tabular}{@{}c|ccc@{}}
    \toprule
           & $|\cos \langle \hat{\mathbf{w}}_\mathsf{b}, \mathbf{w}_\mathsf{b} \rangle|$ \(\uparrow\)             & $|\cos \langle \hat{\mathbf{w}}_\mathsf{b}, \mathbf{w}_\mathsf{t} \rangle|$ \(\downarrow\)     & $\Delta \cos \uparrow$       \\ \midrule
    CelebA & 0.08$\pm$0.02 & 0.004$\pm$0.002 & 0.07$\pm$0.02 \\
    FFHQ   & 0.08$\pm$0.05 & 0.01$\pm$0.005 & 0.07$\pm$0.04 \\ \bottomrule
    \end{tabular}
    \caption{Results of different random initializations of biased attribute hyperplane on face images. The results on two datasets (CelebA and FFHQ) are averaged over three random seeds. The generator is StyleGAN pretrained on FFHQ.}
    \label{tab.sensitivity_to_init_face}
\end{table}

\section{Ablation Studies}
\label{supp.sec.ablation_study}

\begin{table*}[h]
    \centering
    \begin{tabular}{c|c|ccccc}
    \toprule
    distribution & method                        & $|\cos \langle \hat{\mathbf{w}}_\mathsf{b}, \mathbf{w}_\mathsf{b} \rangle|$ \(\uparrow\) & $|\cos \langle \hat{\mathbf{w}}_\mathsf{b}, \mathbf{w}_\mathsf{t} \rangle|$ \(\downarrow\) & $\Delta \cos$ \(\uparrow\) & \%leading \(\uparrow\) & TV       \\ \midrule
    \multirow{3}{*}{balanced} & VAE-based      & 0.21\(\pm\)0.21            & 0.16\(\pm\)0.13            & 0.05$\pm$0.20                & 16.67\%       &  0.15$\pm$0.07  \\
    & Hessian Penalty              & \textbf{0.24\(\pm\)0.16}   & 0.26\(\pm\)0.16            & -0.02$\pm$0.24                & 31.67\%      & 0.14$\pm$0.05  \\
    & \textbf{Ours}                       & 0.23\(\pm\)0.18          & \textbf{0.10\(\pm\)0.11}   & \textbf{0.12$\pm$0.21}      & \textbf{51.67\%} & 0.13$\pm$0.05 \\ \midrule
    \multirow{3}{*}{skewed} & VAE-based         & 0.17\(\pm\)0.18            & 0.19\(\pm\)0.16   & -0.01$\pm$0.22                & 28.33\% & 0.10$\pm$0.04 \\
    & Hessian Penalty     & 0.24\(\pm\)0.17  & 0.29\(\pm\)0.18            & -0.04$\pm$0.23           & 11.67\%     & 0.09$\pm$0.04    \\
    & \textbf{Ours}        & \textbf{0.24\(\pm\)0.16}            & \textbf{0.13\(\pm\)0.11}   & \textbf{0.11$\pm$0.17}       & \textbf{60.00\%} & 0.09$\pm$0.04 \\ \bottomrule
    \end{tabular}
    \caption{Ablation study of data distribution of training data for generative models on SmallNORB~\cite{lecun2004IEEEConf.Comput.Vis.PatternRecognit.CVPRa} dataset. ``Balanced'' means that the distribution of generative model's training data is balanced and ``skew'' denotes that the distribution of generative model's training data is skewed (\textit{same} skewness with the training data of the classifier). For the generative model whose training set is skewed, the TV results of the ground-truth biased attribute and target attribute are 0.02$\pm$0.02, 0.03$\pm$0.02, respectively.}
    \label{tab.balance_skewed}
\end{table*}

\subsection{Ablation Study on Skewed Dataset for Training Generative Models}
\label{sec.ablate_gen_skew}
In the experiments on the disentanglement datasets, the distribution of the generative model's training set is balanced, meaning that the distribution of the target attribute and the distribution of the biased attribute are independent of each other. This setting may not be feasible when such a balanced training set is unavailable. To study the performance of our method and the baseline methods when the balanced training data is unavailable, we train the generative models with the \textit{same} skewed distribution (\ie, $S=0.9$) in the training set for training the biased classifier (see Sec.~\ref{subsec.skewed_train_cls} in this supplementary material). To obtain accurate ground-truth for evaluation, we still use the balanced dataset to compute the ground-truth hyperplanes (method for computing the ground-truth hyperplanes is shown in Sec.~\ref{subsec.gt_hyperplane}). The results on SmallNORB~\cite{lecun2004IEEEConf.Comput.Vis.PatternRecognit.CVPRa} dataset are shown in Tab.~\ref{tab.balance_skewed}. Our method outperforms baseline methods in all metrics with the generative models trained on training data in the skewed distribution. While baseline methods' performances become worse when using the generative models trained on skewed data, our method can still maintain consistent performances. The results demonstrate that baseline methods perform worse when the training data is in a skewed distribution, and our method can better discover the biased attribute and achieve better disentanglement \wrt the target attribute.

\begin{table*}[h]
    \centering
    \begin{tabular}{c|cccc}
    \toprule
          & $|\cos \langle \hat{\mathbf{w}}_\mathsf{b}, \mathbf{w}_\mathsf{b} \rangle|$ \(\uparrow\) & $|\cos \langle \hat{\mathbf{w}}_\mathsf{b}, \mathbf{w}_\mathsf{t} \rangle|$ \(\downarrow\) & $\Delta \cos$ $\uparrow$ & TV \\ \midrule
    $K = \varnothing$ & \textbf{0.26$\pm$0.17}             & 0.12$\pm$0.12       & \textbf{0.14$\pm$0.21}     & 0.13$\pm$0.04                \\
    $K \neq \varnothing$         & 0.23$\pm$0.18             & \textbf{0.10$\pm$0.11} & 0.11$\pm$0.22 & 0.13$\pm$0.05 \\ \bottomrule
    \end{tabular}
    \caption{Ablation study on the known attributes in the \textit{orthogonalization penalty} on SmallNORB~\cite{lecun2004IEEEConf.Comput.Vis.PatternRecognit.CVPRa} dataset. $K = \varnothing$ means that the set of known attributes $K$ is not used in \textit{orthogonalization penalty} and only the target attribute is used. $K \neq \varnothing$ denotes that all of known attributes and the target attribute are used in \textit{orthogonalization penalty}.}
    \label{tab.known_attributes}
\end{table*}

\subsection{Ablation Study on Known Attributes}
In the \textit{orthogonalization penalty}, users can provide a set of known attributes $K$ to exclude the case that the discovered biased attribute is identical with one of the known attributes. Here we also show the results on SmallNORB~\cite{lecun2004IEEEConf.Comput.Vis.PatternRecognit.CVPRa} dataset when known attributes are not provided (\ie, $K=\varnothing$) in Tab~\ref{tab.known_attributes}. The results show that known attributes make performance in $|\cos \langle \hat{\mathbf{w}}_\mathsf{b}, \mathbf{w}_\mathsf{b} \rangle|$ slightly worse and improve the performance in  $|\cos \langle \hat{\mathbf{w}}_\mathsf{b}, \mathbf{w}_\mathsf{t} \rangle|$. We regard that the known attributes introduce a stronger constraint for discovering the biased attribute, but they are still helpful for better disentanglement \wrt the target attribute.

\begin{table*}[h]
    \centering
    \begin{tabular}{@{}c|c|ccccc@{}}
    \toprule
    skewness                & method          & $|\cos \langle \hat{\mathbf{w}}_\mathsf{b}, \mathbf{w}_\mathsf{b} \rangle|$ \(\uparrow\) & $|\cos \langle \hat{\mathbf{w}}_\mathsf{b}, \mathbf{w}_\mathsf{t} \rangle|$ \(\downarrow\) & $\Delta \cos \uparrow$ & \%leading $\uparrow$ & TV \\ \midrule
    \multirow{3}{*}{$S=0.99$} & VAE-based    & 0.22$\pm$0.23             & 0.16$\pm$0.14   & 0.06$\pm$0.22  & 10.00\%  & 0.14$\pm$0.06      \\
                            & Hessian Penalty & 0.21$\pm$0.17             & 0.23$\pm$0.16  & -0.01$\pm$0.25 & 23.33\%  & 0.13$\pm$0.04        \\
                            & \textbf{Ours}            & \textbf{0.25$\pm$0.19}   & \textbf{0.11$\pm$0.11}  & \textbf{0.14$\pm$0.22} & \textbf{48.33\%}   & 0.12$\pm$0.04       \\ \midrule
    \multirow{3}{*}{$S=0.95$} & VAE-based    & 0.20$\pm$0.22             & 0.15$\pm$0.13    & 0.05$\pm$0.20  & 13.33\%  & 0.14$\pm$0.06     \\
                            & Hessian Penalty & \textbf{0.25$\pm$0.17}             & 0.20$\pm$0.16 & 0.05$\pm$0.24  & 35.00\%  & 0.13$\pm$0.04        \\
                            & \textbf{Ours}            & 0.23$\pm$0.18             & \textbf{0.10$\pm$0.10} & \textbf{0.13$\pm$0.21} & \textbf{51.67\%}  & 0.12$\pm$0.04         \\ \midrule
    \multirow{3}{*}{$S=0.9$}  & VAE-based    & 0.21$\pm$0.21             & 0.16$\pm$0.13     & 0.05$\pm$0.02  & 16.67\% & 0.15$\pm$0.07     \\
                            & Hessian Penalty & \textbf{0.24$\pm$0.16}   & 0.26$\pm$0.16  & -0.02$\pm$0.25  & 31.67\%  & 0.14$\pm$0.05       \\
                            & \textbf{Ours}            & 0.23$\pm$0.18             & \textbf{0.10$\pm$0.11}  & \textbf{0.12$\pm$0.21} & \textbf{51.67\%}  & 0.13$\pm$0.05 \\ \midrule
    \multirow{3}{*}{$S=0.75$} & VAE-based    & 0.20$\pm$0.22             & 0.15$\pm$0.13   & 0.05$\pm$0.21   & 13.33\%  & 0.15$\pm$0.07     \\
                            & Hessian Penalty & \textbf{0.23$\pm$0.17}             & 0.19$\pm$0.15  & 0.03$\pm$0.25 & 40.00\% & 0.14$\pm$0.05         \\
                            & \textbf{Ours}     & \textbf{0.23$\pm$0.18}             & \textbf{0.11$\pm$0.10}  & \textbf{0.12$\pm$0.21} & \textbf{46.67\%}  & 0.13$\pm$0.04        \\ \bottomrule
    \end{tabular}
    \caption{Ablation study on the skewness of distribution of data for training the classifiers on SmallNORB~\cite{lecun2004IEEEConf.Comput.Vis.PatternRecognit.CVPRa} dataset.}
    \label{tab.skewness_cls}
\end{table*}

\subsection{Ablation Study on Skewness of Distribution of Data for Training Classifiers}
\label{subsec.ablate_skewness}
We conduct the ablation study on the distribution of data for training the classifiers on SmallNORB~\cite{lecun2004IEEEConf.Comput.Vis.PatternRecognit.CVPRa} dataset. As results shown in Tab.~\ref{tab.skewness_cls}, higher skewness (\ie, $S$) (introduced in Sec.~\ref{subsec.skewed_train_cls}) leads to better results in $|\cos \langle \hat{\mathbf{w}}_\mathsf{b}, \mathbf{w}_\mathsf{b} \rangle|$, indicating that the higher skewness makes it easier for discovering the biased attribute. The results show that our method can beat baseline methods in both metrics under all skewness settings.

\section{Ablation Study on Differnt Random Initializations of Biased Attribute Hyperplane}
To investigate whether our method is sensitive to different random initialization of biased attribute hyperplanes, we conduct experiments with three different random seeds on face images with the same setting introduced in Sec.~\ref{subsec.exp_celeba}. The mean and standard deviation of results over different random seeds are reported in Tab.~\ref{tab.sensitivity_to_init_face}, which shows that our method is robust to different random initializations of biased attribute hyperplane.

\begin{table}[t]
    \centering
    \begin{adjustbox}{width=\columnwidth,center}
    \begin{tabular}{@{}c|c|cccc@{}}
    \toprule
                        & method          & $|\cos \langle \hat{\mathbf{w}}_\mathsf{b}, \mathbf{w}_\mathsf{b} \rangle|$ \(\uparrow\)             & $|\cos \langle \hat{\mathbf{w}}_\mathsf{b}, \mathbf{w}_\mathsf{t} \rangle|$ \(\downarrow\)     & $\Delta \cos \uparrow$ & TV    \\ \midrule
    \parbox[t]{1mm}{\multirow{3}{*}{\rotatebox[origin=c]{90}{ {\scriptsize SmallNORB} }}}  & VAE-based       & 0.08          & 0.19          & -0.12         & 0.16   \\
                               & $\mathcal{L}_H$ & \textbf{0.14} & 0.33          & -0.19         & 0.18  \\
                               & Ours            & 0.13          & \textbf{0.07} & \textbf{0.06} & 0.13 \\ \midrule
    \parbox[t]{1mm}{\multirow{3}{*}{\rotatebox[origin=c]{90}{ {\footnotesize dSprites} }}}   & VAE-based       & \textbf{0.08} & 0.17          & -0.10         & 0.05  \\
                               & $\mathcal{L}_H$ & \textbf{0.08} & 0.53          & -0.45         & 0.05        \\
                               & Ours            & 0.05          & \textbf{0.04} & \textbf{0.01} & 0.01  \\ \bottomrule
    \end{tabular}
    \end{adjustbox}
    \caption{Quantitative results of qualitative results shown in Fig.~\ref{fig.disentanglement_datasets}.}
    \label{tab.quantitative_fig_3}
\end{table}

\begin{table*}[t]
    \centering
    \begin{tabular}{@{}c|c|c|cccc@{}}
    \toprule
    classifier & StyleGAN  & method   & $|\cos \langle \hat{\mathbf{w}}_\mathsf{b}, \mathbf{w}_\mathsf{b} \rangle|$ \(\uparrow\)             & $|\cos \langle \hat{\mathbf{w}}_\mathsf{b}, \mathbf{w}_\mathsf{t} \rangle|$ \(\downarrow\)     & $\Delta \cos \uparrow$       & TV         \\ \midrule
    \multirow{4}{*}{CelebA}      & \multirow{2}{*}{CelebA-HQ} & $\mathcal{L}_H$ & 0.0007        & 0.02              & -0.02         & 0.007         \\
                                 &                           & Ours            & \textbf{0.07} & \textbf{0.001}    & \textbf{0.07} & \textbf{0.11}  \\ \cmidrule(l){2-7}
                                 & \multirow{2}{*}{FFHQ}     & $\mathcal{L}_H$ & 0.04          & 0.02              & 0.02          & 0.03          \\
                                 &                           & Ours            & \textbf{0.05} & \textbf{0.003}    & \textbf{0.05} & \textbf{0.12} \\ \midrule
    \multirow{4}{*}{FFHQ}        & \multirow{2}{*}{CelebA-HQ} & $\mathcal{L}_H$ & 0.03          & 0.007             & 0.03          & 0.05          \\
                                 &                           & Ours            & \textbf{0.28} & $\mathbf{10^{-5}}$ & \textbf{0.28} & \textbf{0.11} \\ \cmidrule(l){2-7}
                                 & \multirow{2}{*}{FFHQ}     & $\mathcal{L}_H$ & \textbf{0.06} & 0.02              & 0.03          & 0.05          \\
                                 &                           & Ours            & \textbf{0.06} & \textbf{0.004}    & \textbf{0.06} & \textbf{0.11} \\ \bottomrule
    \end{tabular}
    \caption{Detailed results of each experiment setting on face images. The first two columns denote the training datasets of the target attribute classifier and StyleGAN, respectively. Our method beats the baseline method under every experiment setting.}
    \label{tab.detailed_face}
\end{table*}

\section{Detailed Experimental Results}
\label{sec.more_detailed_exp_results}

\paragraph{Quantitative Results in Fig.~\ref{fig.disentanglement_datasets}} We show the quantitative results of the Fig.~\ref{fig.disentanglement_datasets} in Tab.~\ref{tab.quantitative_fig_3}. The results prove that our method can more accurately predict the biased attribute and keep a better disentanglement \wrt the target attribute, which is also reflected in Fig.~\ref{fig.disentanglement_datasets}.

\paragraph{Detailed Results on Face Image Datasets} We show the detailed results of experiments on face image datasets in Tab.~\ref{tab.detailed_face}. Our method achieves better results in all experimental settings than the Hessian Penalty method. Especially, our method can achieve much better results when using the StyleGAN trained on CelebA-HQ to discover the biased attribute in the classifier trained on FFHQ (see 6th row in Tab.~\ref{tab.detailed_face}). We suspect that the biased attribute and the target attribute are less correlated in CelebA-HQ than FFHQ, making it easier for discovering the biased attribute in the classifier trained on FFHQ dataset. We also include a discussion on this in Sec.~\ref{sec.discuss_orth_penalty} and Sec.~\ref{sec.limitation_future_works}.

\begin{figure*}[t]
    \centering
    \includegraphics[width=\linewidth]{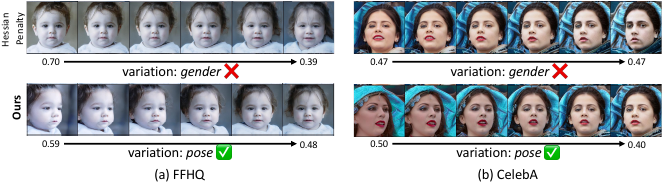}
    \caption{Additional qualitative comparison of the traversal images of predicted biased attributes synthesized by StyleGAN~\cite{karras2019IEEEConf.Comput.Vis.PatternRecognit.CVPR} trained on FFHQ~\cite{karras2019IEEEConf.Comput.Vis.PatternRecognit.CVPR} dataset. Numbers below the image is the predicted probability of ``male'' from the \textit{gender} classifier. The training datasets of the target attribute classifier of (a) and (b) are FFHQ and CelebA, respectively. The Hessian Penalty method predicts \textit{gender} as the biased attribute, which is a trivial solution for a \textit{gender} classifier. In (b), although the traversal images from Hessian Penalty also vary in terms of the \textit{skin tone} attribute, it has very little variation in terms of the classifier's prediction (\ie, 0.47 to 0.47). In comparison, our method can correctly predict the pseudo-ground-truth biased attribute \textit{pose} on two datasets.}
    \label{fig.supp_target_only}
\end{figure*}

\begin{figure}[t]
    \centering
    \includegraphics[width=\linewidth]{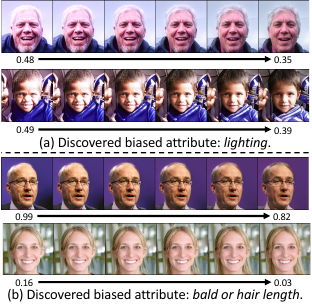}
    \caption{Additional qualitative results on discovered biased attributes by setting the set of known attributes $K$ to all considered attributes generated by StyleGAN~\cite{karras2019IEEEConf.Comput.Vis.PatternRecognit.CVPR} trained on FFHQ~\cite{karras2019IEEEConf.Comput.Vis.PatternRecognit.CVPR} dataset. The target attributes of (a) and (b) is \textit{gender}. The target attribute classifiers in (a) and (b) are trained on CelebA and FFHQ, respectively. Our method can successfully discover \textit{lighting} and \textit{bald or hair length} biased attributes.}
    \label{fig.supp_orth_all}
\end{figure}

\section{Qualitative Results}
Here we show more qualitative results on face images and images from other domains.

\subsection{Qualitative Comparisons on Face Images}
\label{supp.subsec.faces_qualitative_comparison}
We show more qualitative comparisons on face images in Fig.~\ref{fig.supp_target_only}. Compared with Hessian Penalty~\cite{peebles2020Eur.Conf.Comput.Vis.ECCV}, our method can accurately discover the biased attribute \textit{pose} while keeping the disentanglement \wrt the target attribute \textit{gender}. In contrast, the Hessian Penalty~\cite{peebles2020Eur.Conf.Comput.Vis.ECCV} method predicts target attribute \textit{gender} in Fig.~\ref{fig.supp_target_only} (a) and Fig.~\ref{fig.supp_target_only} (b), which are trivial solutions.

\subsection{Discovering Other Biased Attributes on Face Images}
By setting all considered attributes as known attributes in the \textit{orthogonalization penalty}, our method can discover biased attributes other than the known attributes. We additionally show more results in Fig.~\ref{fig.supp_orth_all}. Our method can discover \textit{lighting} and \textit{bald or hair length} biased attributes, which are not the known attributes. We found that the male images have variations in terms of the \textit{bald} attribute, and female images have variations in terms of the \textit{hair length} attribute based on the very same predicted biased attribute hyperplane. Since \textit{bald} and \textit{hair length} are all closely related, we merge them together and regard it as \textit{bald or hair length} attribute. We also admit that our method cannot achieve perfect disentanglement with other attributes such as \textit{beard} in the first row of Fig.~\ref{fig.supp_orth_all} (a). We will discuss this problem in Sec.~\ref{sec.limitation_future_works}.

\subsection{Images from Other Domains}
\begin{figure*}[t]
    \centering
    \includegraphics[width=0.85\linewidth]{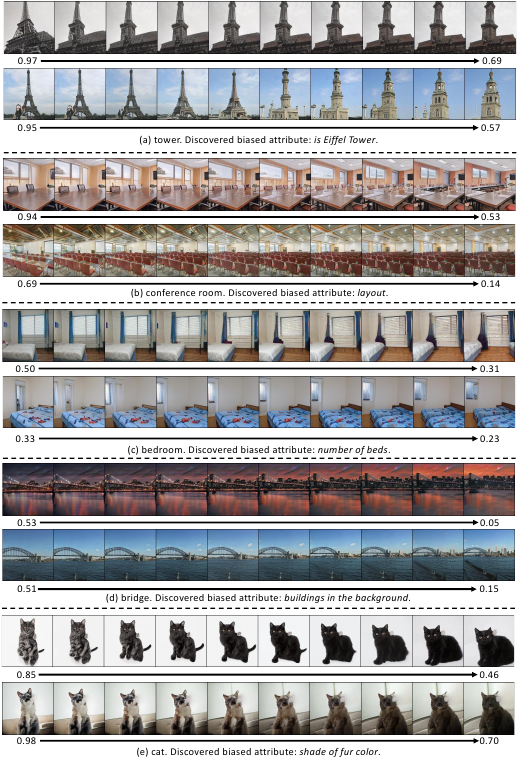}
    \caption{Additional qualitative results on the discovered biased attribute of classifiers for classifying \textit{tower}, \textit{conference room}, \textit{bedroom}, \textit{bridge}, and \textit{cat} images. Numbers below images are predicted probability by the classifier.}
    \label{fig.supp_other_domain}
\end{figure*}

We additionally show more qualitative results on the discovered biased attribute for classifiers on images from other domains in Fig.~\ref{fig.supp_other_domain}. Our method can successfully discover unnoticeable biased attributes such as \textit{is Eiffel Tower}, \textit{layout}, \textit{number of beds}, \textit{buildings in the background}, \textit{shade of fur color} for tower, conference room, bedroom, bridge, and cat classifiers, respectively.

\begin{figure*}[t]
    \centering
    \includegraphics[width=\linewidth]{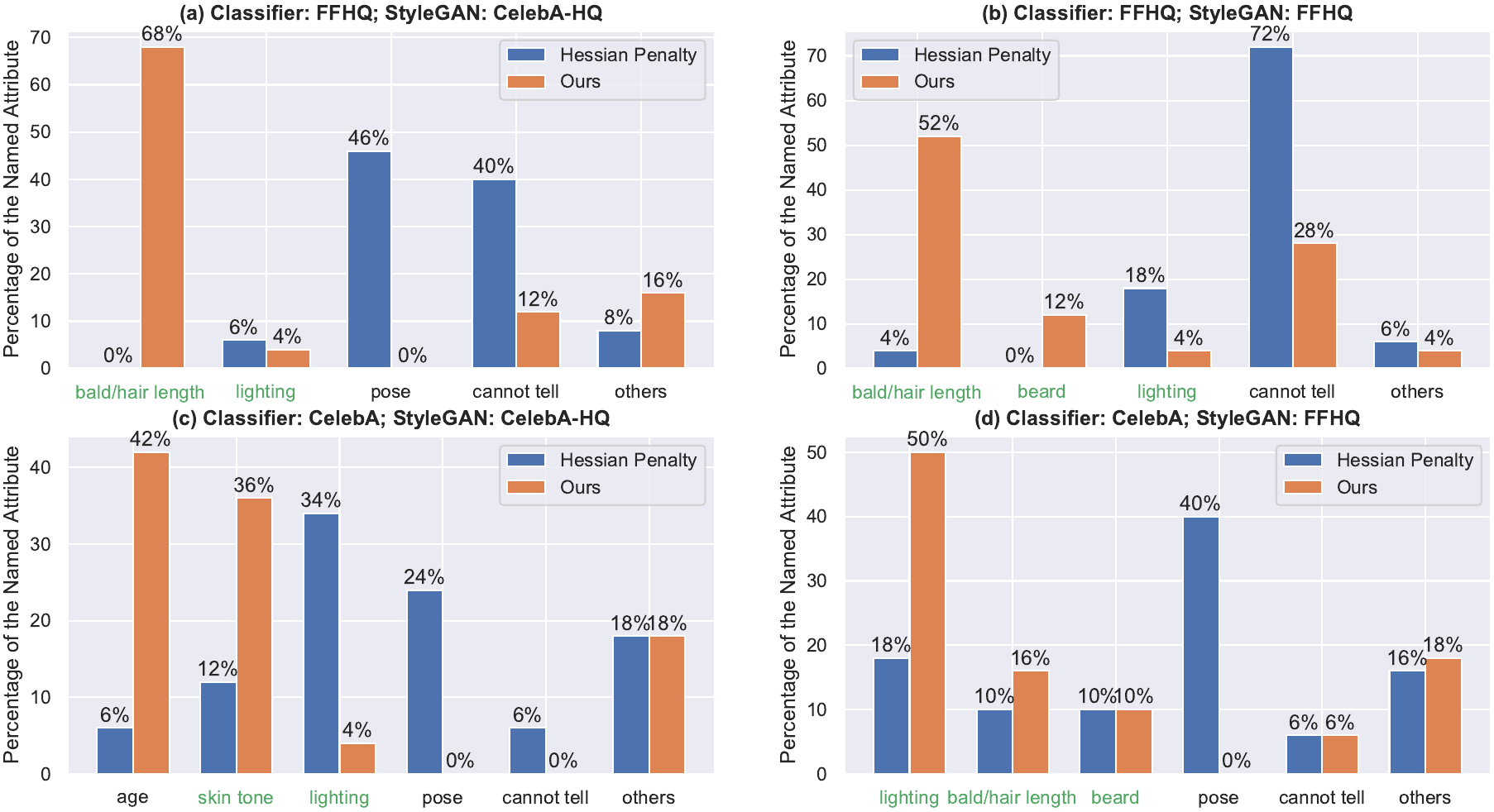}
    \caption{User study on face images. Four bar charts correspond to four experiment settings on face images. The title of each chart denotes the experiment setting. For example, ``Classifier: FFHQ; StyleGAN: CelebA-HQ'' means that the target attribute classifier is trained on FFHQ, and the StyleGAN is trained on CelebA-HQ. In each bar chart, the x-axis is the attribute named by the users. The y-axis is the percentage of the attribute out of all named attributes by users. A higher percentage means that users agree more on that attribute. The attributes in \textcolor{green}{green} are out of the known-biased attribute set $K$. Attributes in green with higher percentages and the attributes in black with lower percentages mean that the method can better find the unknown attributes. In experiment settings (a), (b), and (d), all users agree that our method find the biased attribute \textit{bald/hair length}, whereas the Hessian Penalty method can only find known-biased attribute \textit{pose} (in (a), (d)), or users cannot tell the attribute (see ``cannot tell'' in (b)). In the experiment setting (c), although the top-1 named attribute is a known attribute \textit{age}, the named attribute with the second largest percentage is \textit{skin tone}, whose percentage is still larger than the percentage of the top-1 attribute \textit{lighting} from the Hessian Penalty method. We suspect the reason is that our method does not achieve perfect disentanglement between \textit{age} and \textit{skin tone}. In conclusion, our method can better find other unknown biased attributes than the Hessian Penalty method in all experiment settings on face images.}
    \label{fig.user_study_faces}
\end{figure*}

\begin{figure*}[t]
    \centering
    \includegraphics[width=\linewidth]{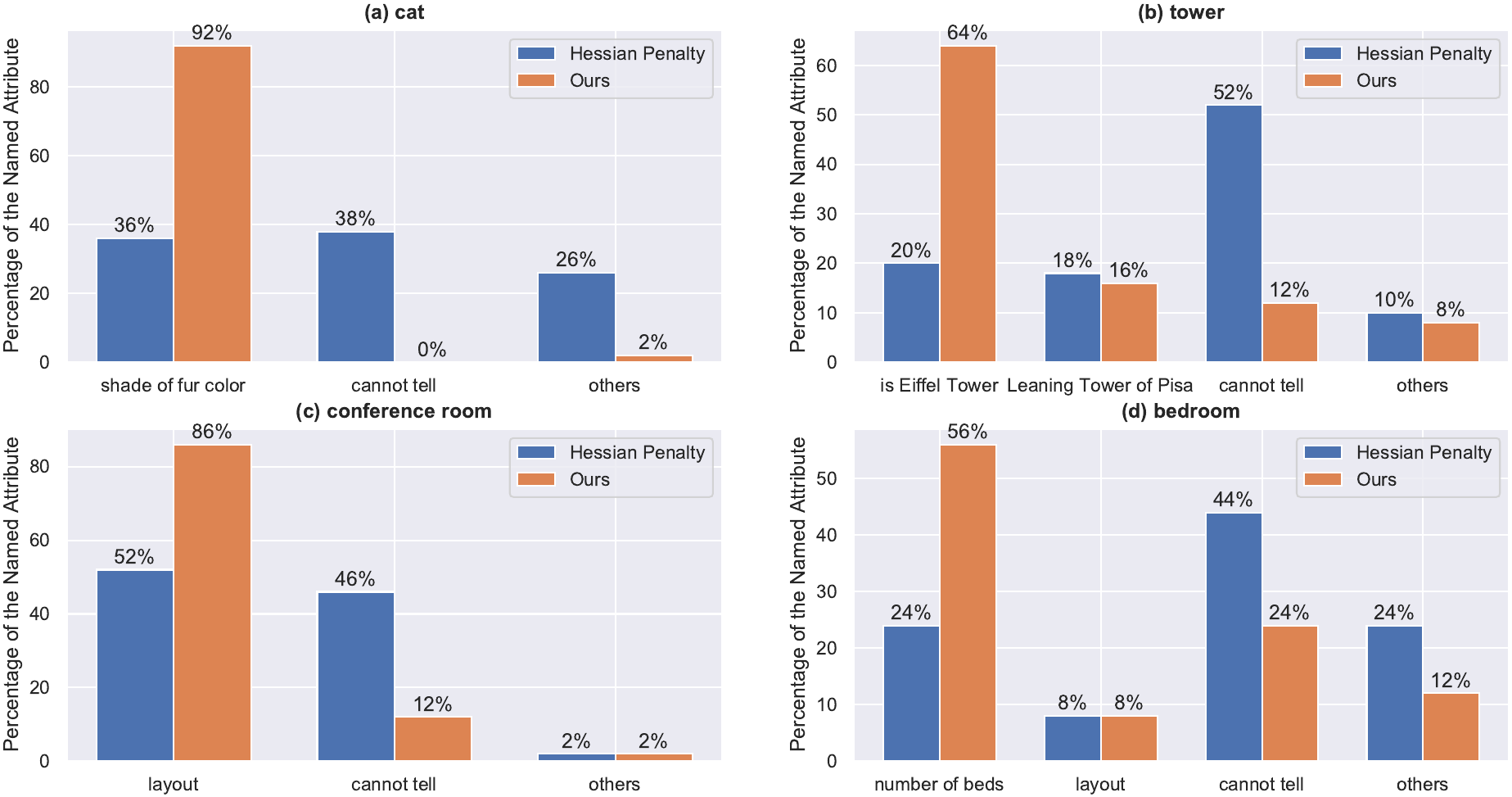}
    \caption{User study on other domains of images. Four bar charts correspond to four image domains (see titles of each chart). In each bar chart, the x-axis is the attribute named by the users. The y-axis is the percentage of the attribute out of all named attributes by users. A higher percentage means that users agree more on that attribute. The user study results show that the biased attributes predicted by the Hessian Penalty method are either uniformly distributed or cannot be told by the users (\eg, ``cannot tell'' in (b) and (d)). In contrast, the biased attributes from our method are more concentrated on the one attribute (\eg, \textit{shade of fur color}). In conclusion, our method can better find the unknown biased attributes in diverse domains of images than the Hessian Penalty method.}
    \label{fig.user_study_other_domains}
    \vspace{-2mm}
\end{figure*}

\section{User Study}
\label{supp.sec.user_study}
We conduct the user study to verify that our method can find unknown-biased attributes that are difficult for the baseline method on more images comparable to Fig.~\ref{fig.celeba_all}, \ref{fig.other_domains}. Ten subjects are asked to name the attribute of traversal images synthesized by $\mathcal{L}_H$ and our method without knowing the images are generated by which method. The user can also say ``cannot tell'' the attribute from the traversal image (denoted as ``cannot tell'' in Fig.~\ref{fig.user_study_faces} and Fig.~\ref{fig.user_study_other_domains}) when the users find it hard to interpret the variation of the traversal images. If the user regards multiple attributes in the traversal images, we ask the users to name the most salient one. For a fair comparison, we use the \textit{same} sampled latent vector $\mathbf{z}$ for two methods to synthesize traversal images (40 face traversal images, 40 other-domain traversal images). To further let the Hessian Penalty method find other biased attributes, we remove $|K|$ (cardinality of set $K$) predicted hyperplanes which are top-$|K|$ similar (\ie, high absolute value of cosine similarity) with the known-biased attributes $K$, which is a similar procedure as we introduce how to adapt the baseline method in Sec.~\ref{sec.baseline_method}. After collecting the user study results, we compute the percentage of each attribute named by the user. For example, ``$68\%$'' of the \textit{bald / hair length} attribute of our method in Fig.~\ref{fig.user_study_faces} (a) means that among all the named attributes on the traversal images generated in the experiment setting (a), $68\%$ of them are \textit{bald / hair length} attributes. All other attributes that are rarely named by the users are merged into ``others'' in Fig.~\ref{fig.user_study_faces} and Fig.~\ref{fig.user_study_other_domains}.

\paragraph{Discovered Other Biased Attributes on Face Images}
The results of the user study on finding other biased attributes on face image datasets are shown in Fig.~\ref{fig.user_study_faces}. For the Hessian Penalty method, by looking at the largest percentage among all attributes, most users agree that it still predicts the known biased attribute $\textit{pose}$ in experiment settings (a) and (d), and the users cannot tell the attribute in experiment setting (b). Although the Hessian Penalty can predict one unknown biased attribute \textit{lighting} in experiment setting (c), its percentage ($34\%$) is still lower than the percentage of \textit{skin tone} attribute ($36\%$) predicted by our method. In the other three experiment settings (a), (b), and (d), by looking at the largest percentage among all attributes, most users agree that our method predicts \textit{bald / hair length} and \textit{lighting} biased attributes. In conclusion, our method can find the biased attributes that are difficult for the baseline method.

\paragraph{Discovered Biased Attributes on Images from Other Domains}
We conduct the user study on four categories of images: cat, tower, conference room, and bedroom. The results of the user study on discovered biased attributes on images from other domains are shown in Fig.~\ref{fig.user_study_other_domains}. For the Hessian Penalty method, the named attributes are either uniformly distributed (see Hessian Penalty results in Fig.~\ref{fig.user_study_other_domains} (a) (b)), or the user cannot tell the attributes from the traversal images (see Hessian Penalty results in Fig.~\ref{fig.user_study_other_domains} (b) (d)). Hence, it is hard to tell the biased attribute from the traversal images synthesized by the Hessian Penalty's biased attribute prediction. In contrast, by looking at the largest percentage numbers, most users agree that our method can find \textit{shade of fur color}, \textit{is Eiffel Tower}, \textit{layout}, and \textit{number of beds} biased attributes, meaning that users can easily tell the biased attribute from the traversal images generated by our method. In conclusion, our method can better discover the biased attributes in diverse domains of images, which are hard to be found by the baseline method.

\section{Discussion}

\subsection{Why $|\cos \langle \hat{\mathbf{w}}_\mathsf{b}, \mathbf{w}_\mathsf{b} \rangle|$ results on face image datasets are smaller than the ones on disentanglement datasets?} One may question why the $|\cos \langle \hat{\mathbf{w}}_\mathsf{b}, \mathbf{w}_\mathsf{b} \rangle|$ results on the faces image datasets (Tab.~\ref{tab.celeba}) are smaller than the results on disentanglement datasets (Tab.~\ref{tab.disentanglement_datasets_results}). We suspect two reasons for it. First, the numbers of latent space dimensions used in the two experiments are different (512 vs. 10). It would be more difficult for face image experiments to optimize in latent space with larger dimensions (512). Second, discovering the biased attribute on face image datasets is harder than identifying biases on disentanglement datasets because the former datasets are in-the-wild datasets, whereas the latter synthetic datasets only contain finite sets of attributes.

\subsection{Why use $\Delta \cos$ as the major evaluation metric?}
One may ask that why we use $\Delta \cos$ as the major evaluation metric. The reason is that our ultimate goal is to let the human interpret the biased attribute from the traversal images (see Fig.~\ref{fig:teaser} (b)). Suppose that the traversal images contain the variations of two attributes: the biased attribute and the target attribute. In this case, although the classifier has large prediction variations among the traversal images, humans still cannot decide which attribute (biased attribute or target attribute) is the real cause for the prediction variations. Therefore, it would be better for humans to make a causal conclusion if the traversal images only contain the variation of the biased attribute and do not contain the variation of the target attribute.

\subsection{``Small'' Variation}
\label{supp.subsec.small_variation}
One may regard the variation of predicted probabilities that do not switch the 0.5-threshold (\eg, 0.48 to 0.35 in Fig.~\ref{fig.supp_orth_all}) are ``small'' for a binary classifier. However, we believe such ``small'' variation is still value for the \textit{unknown bias discovery task} for the following reasons:

First, such variations still break the fairness criterion. Second, we do not think switching 0.5 threshold is necessary because 1) a larger threshold may be required for some safety-critical scenarios; 2) when a classifier needs to give the ranking of different input images (such as image retrieval task) based on the predicted probability, the threshold is not needed, and the biased ranking may still lead to unfairness issue; 3) in a multi-class classification setting (\ie, experiments on other domains of images), the threshold is not 0.5 and such ``small'' variation can alter the classifier's top-1 prediction. Third, such ``small'' variations can still provide insights to dataset curators to mitigate such biases, which could be learned by other networks.

\subsection{Discussions on Orthogonalization Penalty}
\label{sec.discuss_orth_penalty}

One may worry that the \textit{orthogonalization penalty} may prevent the method from finding the biased attribute highly correlated with the target attribute, based on the assumption that the high correlation in the training set of the classifier leads to a high absolute value of cosine similarity between the biased attribute normal vector and the target attribute normal vector in the latent space of the generative model. For example, if the \textit{hair length} biased attribute is highly correlated with the target attribute \textit{gender} in the training set of the classifier (\eg, ``long hair female'' and ``short hair male'' are overrepresented in the dataset), one may think that $\mathcal{L}_\perp$ may prevent our method from finding the \textit{hair length} attribute hyperplane due to its high absolute cosine similarity with \textit{gender} attribute hyperplane. We list our arguments and possible solutions to this question by the following points:

First, $\mathcal{L}_\perp$ is a soft penalty instead of a strict constraint, which still allows the correlation between the predicted biased attribute and the target one. In fact, we tried strict constraint but it is hard to optimize, which fails to predict biased attribute, resulting in low $|\cos \langle \hat{\mathbf{w}}_\mathsf{b}, \mathbf{w}_\mathsf{b} \rangle|$ values. That is also reflected by the ablation study on the soft orthogonalization penalty in Tab.~\ref{tab.ablate_orth_hessian}, where even adding the soft orthogonalization penalty has already decreased the $|\cos \langle \hat{\mathbf{w}}_\mathsf{b}, \mathbf{w}_\mathsf{b} \rangle|$ results. Hence, the proposed soft \textit{orthogonalization penalty} is still better than the strict orthogonalization constraint for solving the aforementioned issue.

Second, the ablation study on the skewness of the generative model's training data (Sec.~\ref{sec.ablate_gen_skew}) has already proved that our method can still accurately predict the biased attribute even if the generative model's training data has the \textit{same} skewness of classifier's training data (see results in Tab.~\ref{tab.balance_skewed}). Note that the performance of two baseline methods drops when using the generative model trained on skewed data, whereas our method still maintains high performance.

Third, we observe that using a more disentangled latent space (\eg, $\mathcal{W}$-space of StyleGAN) of the generative model can also mitigate this issue. We tried using the input noise space ($\mathcal{Z}$-space) of the StyleGAN on face images, which has much lower results than using the $\mathcal{W}$-space of StyleGAN. For example, our method finds the \textit{bald / hair length} biased attribute in the $\mathcal{W}$-space of StyleGAN, and we do not find such attribute in the $\mathcal{Z}$-space. As explained in Sec.~4 of the StyleGAN~\cite{karras2019IEEEConf.Comput.Vis.PatternRecognit.CVPR} paper, compared with the input noise space ($\mathcal{Z}$-space), $\mathcal{W}$-space is a more disentangled latent space, where hyperplanes are less correlated. Hence, using a more disentangled latent space can also address the issue mentioned above.

Fourth, the assumption that attributes are highly correlated in the latent space may not hold when the training datasets of the generative model and the classifier are different. In other words, the biased attribute hyperplane may not be highly correlated with the target attribute hyperplane when the generative's model training data has a weaker skewness between the biased attribute and the target attribute.

Finally, another solution to the issue is using a generative model that trained on only one value of the target attribute. For example, as we did in the experiment on other domains of images (Sec.~\ref{subsec.exp_other_domain}), we only use $\mathcal{L}_V$ and do not use $\mathcal{L}_\perp$ because the target attribute value will not change among the synthesized images (\ie, a \textit{cat} generator will only synthesize \textit{cat} images, and will never synthesize \textit{dog} images.). Although more generators need to be trained for each target attribute value, this could be another solution to the aforementioned problem as no target attribute exists in the generator’s latent space, not to mention the correlation between the biased attribute and the target attribute.

\subsection{Related Methods and Areas}
\label{subsec.related_methods_areas}
The proposed \textit{unknown biased attribute discovery task} can benefit many related methods and areas:

First, many supervised algorithmic de-biasing methods~\cite{wang2020IEEEConf.Comput.Vis.PatternRecognit.CVPRe} require the well-defined biased attribute (or protected attribute) and corresponding labels to mitigate biases. The discovered unknown biases can serve as the definition of the biased attribute, and the corresponding labels can be further collected by humans.

Second, our method can also benefit dataset curation and audition process. The biases in the image classifiers may originate from the training data of the classifier. Therefore, users can balance the distribution of the dataset based on the discovered unknown biases to mitigate the dataset bias~\cite{torralba2011IEEEConf.Comput.Vis.PatternRecognit.CVPR}.

Third, our method also provides a unique perspective for the area of disentanglement methods. As discussed in \cite{locatelloChallengingCommonAssumptions2019}, unsupervised disentanglement is theoretically impossible, and future works should explicitly present the inductive biases and weak supervision used in the framework. From a disentanglement perspective, our method empirically proves that the biases in a down-stream classifier can serve as a weak prior for finding the biased attribute in the latent space of the generative model.

Finally, our work can also give a new research direction to the adversarial attack methods~\cite{goodfellow2015Int.Conf.Learn.Represent.,madryDeepLearningModels2018}. While most adversarial attack methods focus on adding uninterpretable pixel perturbation to the image, our method uses the traversal images to find the interpretable vulnerability of the deep neural networks.

\subsection{Limitations and Future Direction}
\label{sec.limitation_future_works}
We honestly list some limitations of our method.

First, we do not achieve perfect disentanglement on in-the-wild datasets. For example, in Fig.~\ref{fig.supp_orth_all}, the discovered biased attribute \textit{lighting} is entangled with the attribute \textit{beard}. A possible solution is to obtain more hyperplanes of known attributes (\eg, the \textit{beard} attribute) for \textit{orthogonalization penalty}.

Second, we admit that the biased attributes' searching space is decided by the coverage of attributes of the generative model's training data. However, we believe that this will not be a serious problem as long as the users can access either the same training data used by the classifier or even larger and diverse unlabeled datasets to train the generative model.

Lastly, our method only focuses on detecting one biased attribute at a time, while the target attribute could be affected by multiple biased attributes in real-world settings. A possible solution is extending our method by jointly optimizing multiple orthogonalized biased attribute hyperplanes.

The future directions of \textit{unknown biased attribute discovery task} could be tackling our method's aforementioned limitations. In this work, we only study the counterfactual fairness criterion~\cite{kusner2017Adv.NeuralInf.Process.Syst.,joo2020Proc.2ndInt.WorkshopFairnessAccount.Transpar.EthicsMultimed.,denton2019IEEEConf.Comput.Vis.PatternRecognit.CVPRWorkshop,denton2019IEEEConf.Comput.Vis.PatternRecognit.CVPRWorkshopa}. Future works can explore more fairness criteria to define the biased attribute. One interesting direction is to discover unknown biases from object detectors or semantic segmentation networks rather than classifiers.

Furthermore, we base our method on recent advances in generative models that can synthesize photo-realistic images of faces, objects, and simple scenes. However, to the best of our knowledge, no generative models can synthesize photo-realistic images of complex scenes containing various objects and stuff (\eg, images from MS-COCO dataset~\cite{lin2014Eur.Conf.Comput.Vis.ECCV}). Developing methods for discovering unknown biases for models learned from complex scene images is a challenging but valuable research direction for \textit{unknown biased attribute discovery task}.

\end{document}